\documentclass[sigconf, nonacm, table, pdfa]{acmart}
\usepackage{xcolor} 
\usepackage{graphicx}
\usepackage{multicol}
\usepackage{subcaption}
\usepackage{bm}
\usepackage{wrapfig}
\usepackage{makecell}
\usepackage[subtle]{savetrees}
\usepackage{enumitem} 
\usepackage{cleveref}
\usepackage{tikz}
\usepackage{diagbox}
\usepackage{pdfpages}
\usepackage{balance}
\usepackage{hyperref}
\usepackage[a-2b,mathxmp]{pdfx}[2018/12/22]

\newcommand\vldbdoi{10.14778/3648160.3648165}
\newcommand\vldbpages{1214 - 1226}
\newcommand\vldbvolume{17} 
\newcommand\vldbissue{6}
\newcommand\vldbyear{2024}
\newcommand\vldbauthors{\authors} 
\newcommand\vldbtitle{\shorttitle} 
\newcommand\vldbavailabilityurl{github.com/cirquit/hivemind-multi-cloud}
\newcommand\vldbpagestyle{empty} 
\definecolor{diff}{RGB}{0,0,0}
\definecolor{diff1}{RGB}{0,0,0}

\begin{document}

\title[]{How Can We Train Deep Learning Models Across \\Clouds and Continents? An Experimental Study}
\author{Alexander Erben}
\affiliation{%
  \institution{Technical University of Munich}
}
\email{alex.erben@tum.de}
 
\author{Ruben Mayer}
\affiliation{%
  \institution{University of Bayreuth}
}
\email{ruben.mayer@uni-bayreuth.de}

\author{Hans-Arno Jacobsen}
\affiliation{%
  \institution{University of Toronto}
}
\email{jacobsen@eecg.toronto.edu}

\begin{abstract}
This paper aims to answer the question: Can deep learning models be cost-efficiently trained on a global market of spot VMs spanning different data centers and cloud providers?
To provide guidance, we extensively evaluate the cost and throughput implications of training in different zones, continents, and clouds for representative CV, NLP and ASR models.
To expand the current training options further, we compare the scalability potential for hybrid-cloud scenarios by adding cloud resources to on-premise hardware to improve training throughput.
Finally, we show how leveraging spot instance pricing enables a new cost-efficient way to train models with multiple cheap VMs, trumping both more centralized and powerful hardware and even on-demand cloud offerings at competitive prices.
\end{abstract}
\maketitle
\pagestyle{\vldbpagestyle}
\begingroup\small\noindent\raggedright\textbf{PVLDB Reference Format:}\\
\vldbauthors. \vldbtitle. PVLDB, \vldbvolume(\vldbissue): \vldbpages, \vldbyear.\\
\href{https://doi.org/\vldbdoi}{doi:\vldbdoi}
\endgroup
\begingroup
\renewcommand\thefootnote{}\footnote{\noindent
This work is licensed under the Creative Commons BY-NC-ND 4.0 International License. Visit \url{https://creativecommons.org/licenses/by-nc-nd/4.0/} to view a copy of this license. For any use beyond those covered by this license, obtain permission by emailing \href{mailto:info@vldb.org}{info@vldb.org}. Copyright is held by the owner/author(s). Publication rights licensed to the VLDB Endowment. \\
\raggedright Proceedings of the VLDB Endowment, Vol. \vldbvolume, No. \vldbissue\ %
ISSN 2150-8097. \\
\href{https://doi.org/\vldbdoi}{doi:\vldbdoi} \\
}\addtocounter{footnote}{-1}\endgroup

\ifdefempty{\vldbavailabilityurl}{}{
\begingroup\small\noindent\raggedright\textbf{PVLDB Artifact Availability:}\\
The source code, data, and/or other artifacts have been made available at \href{https://github.com/cirquit/hivemind-multi-cloud}{https://github.com/cirquit/hivemind-multi-cloud}.
\endgroup
}
\section{Introduction}
\label{sec:introduction}
Deciding whether to invest in on-premise hardware or move to the cloud for deep learning (DL) is not easy.
Wanting to scale existing infrastructure means paying upfront, as combining cloud and on-premise is not an option with popular DL frameworks due to needing a dedicated high-bandwidth interconnect.
To enabled model- and data-parallelism, current state-of-the-art accelerators have bandwidths of 900~GB/s for intra-node~\cite{elster2022nvidia} and 25~Gb/s for inter-node setups~\cite{sergeev2018horovod,li2020pytorch}. 
\begin{table}[]
\begin{center}
\caption{Average us-west cloud pricing in April '23.}
\vspace*{-2mm}
\scalebox{0.90}{
\begin{tabular}{l|r|r|r}
\backslashbox{\textbf{Type}}{\textbf{Cloud}} & \textbf{GC} & \textbf{AWS} & \textbf{Azure} \\ \hline
T4 Spot      & \cellcolor[HTML]{67F13A}0.180 \$/h      & \cellcolor[HTML]{f1be3a}0.395 \$/h & \cellcolor[HTML]{67F13A}0.134 \$/h \\
T4 On-Demand & \cellcolor[HTML]{f1be3a}0.572 \$/h      & \cellcolor[HTML]{f17d3a}0.802 \$/h & \cellcolor[HTML]{f1be3a}0.489 \$/h \\ \hline
Traffic (inter-zone)           & \cellcolor[HTML]{67F13A}0.01 \$/GB & \cellcolor[HTML]{67F13A}0.01 \$/GB & \cellcolor[HTML]{67F13A}0.00 \$/GB \\
Traffic (inter-region) US      & \cellcolor[HTML]{67F13A}0.01 \$/GB & \cellcolor[HTML]{67F13A}0.01 \$/GB & \cellcolor[HTML]{c3f13a}0.02 \$/GB \\
Traffic (inter-region) EU      & \cellcolor[HTML]{c3f13a}0.02 \$/GB & \cellcolor[HTML]{67F13A}0.01 \$/GB & \cellcolor[HTML]{c3f13a}0.02 \$/GB \\
Traffic (inter-region) ASIA    & \cellcolor[HTML]{f1be3a}0.05 \$/GB & \cellcolor[HTML]{67F13A}0.01 \$/GB & \cellcolor[HTML]{f1be3a}0.08 \$/GB \\
Traffic (inter-region) OCE     & \cellcolor[HTML]{f1be3a}0.08 \$/GB & \cellcolor[HTML]{67F13A}0.01 \$/GB & \cellcolor[HTML]{f1be3a}0.08 \$/GB \\
Traffic                ANY-OCE & \cellcolor[HTML]{f17d3a}0.15 \$/GB & \cellcolor[HTML]{c3f13a}0.02 \$/GB & \cellcolor[HTML]{f1be3a}0.08 \$/GB \\
Traffic (between continents)   & \cellcolor[HTML]{f1be3a}0.08 \$/GB & \cellcolor[HTML]{c3f13a}0.02 \$/GB & \cellcolor[HTML]{c3f13a}0.02 \$/GB 
\end{tabular}
}
\end{center}
\label{tab:cloud-pricing}
\vspace*{-2mm}
\end{table}
Due to the initial investment of the cloud providers in the accelerators, they naturally want to reap profit by maximizing resource utilization.
Therefore, it is common to have "spot" pricing, which offers the VMs at a strongly reduced rate, typically at a 40-90\% discount (\Cref{tab:cloud-pricing}), but with the drawback that the VM can be terminated at any time if another customer is willing to pay the on-demand price~\cite{portella2019statistical}.
Unfortunately, popular DL frameworks have not been developed with failure semantics in mind and cannot adequately deal with peers that fail~\cite{borzunov2022training}.
While services like Amazon Sagemaker~\cite{das2020sagemaker} and projects like Skypilot~\cite{yang2023skypilot} offer automatic job migration in case of VM termination, they are limited to single-node training due to the bandwidth requirements between accelerators.

But what if we could use spot pricing for long-running, distributed jobs and reduce bandwidth requirements to leverage multiple low-cost GPUs?
This could be possible through a framework for collaborative DL training, Hivemind~\cite{hivemind}, which inherently deals with peers that can stop running at any time.
While there is research on how Hivemind can be used for training on spot VMs~\cite{ryabinin2021moshpit,diskin2021distributed,ryabinin2023swarm}, it does not compare the cost-throughput tradeoff for different cloud offerings or perform ablation studies on geographic distribution or model sizes.

\begin{figure} 
    \includegraphics[width=0.49\textwidth]{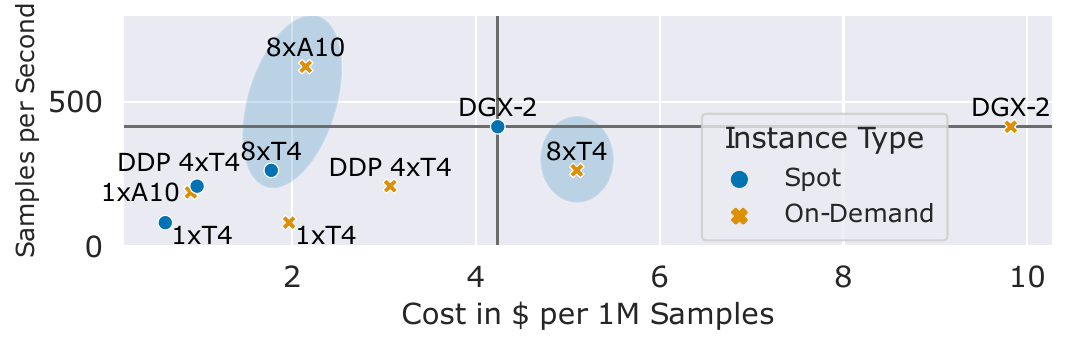}
    \vspace{-7mm} 
    \caption{Cost to throughput tradeoff for ConvNextLarge at different instance types. Our training setups (circled) are cheaper (8xT4) and faster (8xA10) than centralized offerings (DGX-2).} 
    \label{fig:cv-sps-trade-off}
    \vspace*{-3mm} 
\end{figure}

To motivate this new possibility, we trained the ConvNextLarge model~\cite{liu2022convnet} on the Imagenet1K dataset~\cite{deng2009imagenet} on different Google Cloud hardware (T4's and DGX-2), and on the very competitively priced A10 from LambdaLabs (see \Cref{sec:hybrid-cloud-performance} for the full experimental description).
\Cref{fig:cv-sps-trade-off} shows the training throughput and the costs per 1 million processed samples for each setup.
The single node (1xT4, 1xA10, DGX-2) experiments show the current state-of-the-art cost-throughput ratio for training on GC and LambdaLabs.
The DGX-2 node is the fastest, with a throughput of 413~SPS, but it also costs \$6.30/h (\$4.24/1M samples), shown by the horizontal and vertical lines.
The single-accelerator experiments (1xT4, 1xA10) have a better cost-throughput ratio (\$0.62/1M samples and \$0.9/1M samples), but have a much lower throughput of 80 and 185~SPS, respectively.
However, when using our approach of distributing the training between multiple GPUs with Hivemind (circled), we make training possible that is both faster (8xA10, 621~SPS, \$2.15/1M samples) and cheaper (8xT4, 262~SPS, \$1.77/1M samples) than using the DGX-2.
Every cloud provider deals differently with how they price spot instances and network traffic (cf.~\Cref{tab:cloud-pricing}) and has varying interruption rates for different accelerators~\cite{lee2017deepspotcloud}.
Being able to choose the best option was not possible before, and having the option to combine older, more available GPUs is a net benefit for both consumers and cloud providers alike.
 
We aim to develop guidelines and help practitioners assess under which conditions they can cost-efficiently speed up their training tasks with spot instances.
To be able to do this, they need a precise definition of the model size at which geo-distributed spot training becomes viable, what hardware can be used for it, and what the minimum bandwidth and latency are.
We close this research gap by performing a comprehensive analysis of multiple DL tasks from CV and NLP, breaking down how time is spent in each epoch, and comparing them to non-distributed runs to quantify the advantages and disadvantages of distributed spot training.
We determine which models scale with additional spot instances and which cannot be scaled without running into a communication bottleneck or resource inefficiencies.
To quantify total training cost, we assess cost-effectiveness and evaluate a hybrid or multi-cloud approach with popular cloud providers through training on up to four continents.
For comparison of the models' scalability and to show which of them can be trained in a distributed fashion, we introduce the \textit{granularity metric}, the ratio of calculation to communication time, and show how it can be used for predicting performance with different hardware setups.
Finally, we summarize our lessons on how to design geo-distributed spot training and what to watch out for when evaluating the feasibility of such a training regime. Our contributions are:
\hyphenation{dis-tri-bu-ted}
\begin{enumerate}
    \item \textbf{We analyze the impact of multi-cloud training with spot and on-demand instances from Google Cloud (GC), Microsoft Azure, Amazon Web Services (AWS), and LambdaLabs on cost-efficiency.} While we find performance penalties due to remote versus on-premise compute resources, the throughput still scales with increased computing power. By leveraging multiple spot instances with one T4 GPU each, we can be more cost-efficient than a DGX-2 node or the very competitively priced A10 offerings from LambdaLabs.
    \item \textbf{We investigate the suitability of geo-distributed training for various CV and NLP models and hardware configurations on up to four continents.} Not surprisingly, the more parallelizable and the larger the task, the better the performance. Moreover, we verify the scalability claims of the related work and define additional constraints, such as the minimum granularity for effective training. This enables, for the first time, distributed training of smaller million-parameter models (12M-560M) over <1~Gb/s bandwidth and >150ms latency networks.
    \item \textbf{We evaluate two different hybrid-cloud experimental setups with consumer- and server-grade on-premise hardware} and try to improve the throughput with a bandwidth of, at worst, 50~Mb/s to the cloud resources. While we show that it is possible to improve throughput even at these constraints, local cloud offerings are better suited for models that show limited suitability for distributed training.
    \item We summarize our findings of training in a geo-distributed, multi-cloud environment. \textbf{We propose the granularity metric to compare model suitability for distributed spot training} and estimate training performance with additional spot VMs. This provides guidance on the trade-off between performance and cost when using geo-distributed spot instances. {\color{diff}To apply our findings, we perform a case-study on a state-of-the-art model from the ASR domain and achieve speedups on low-end hardware.}
\end{enumerate}

\section{Deep Learning On Spot Instances}
\label{sec:distributed-dl}
In this section, we describe how the Hivemind framework works and how it can enable distributed spot training.
\subsection{Hivemind}
\label{ssec:hivemind}

Hivemind~\cite{hivemind} is a PyTorch-based~\cite{paszke2019pytorch} framework developed initially to enable collaborative DL training where participants could donate their heterogeneous hardware to train a single model together in a data-parallel fashion.
Its main difference to other state-of-the-art distributed training frameworks, such as PyTorch DDP~\cite{li2020pytorch} and DeepSpeed~\cite{rasley2020deepspeed}, is that it runs in a decentralized fashion and can handle peers that drop out at any stage of the training. 
It does so with two features: a distributed hash table~\cite{maymounkov2002kademlia} (DHT) which spans over all participating peers for metadata storage, such as training progress and peer health, and a gradient averaging algorithm that is designed to reduce the impact of lost gradients.
A key difference to other distributed training frameworks is the definition of a \textit{hivemind epoch}, which is the number of samples that must be aggregated before an averaging step is performed.
This sample count is called the \textit{target batch size} (TBS), which corresponds to the minibatch size in standard DL training.
The DHT is used for coordination, and shortly before the TBS is predicted to be reached, the peers start to form the initial groups for averaging.
The time allocated for group forming is called \textit{matchmaking time} and typically runs asynchronously to the training (cf.~\Cref{sec:model-suitability}).
The individual peer gradients are accumulated locally and sent to the other peers via an adaptive all-reduce algorithm (MoshpitSGD~\cite{ryabinin2021moshpit}).
The next hivemind epoch starts after each peer applies the accumulated gradients to the local model.
The advantage of Hivemind for geo-distributed training comes from cumulating different techniques, such as Delayed Parameter Updates~\cite{ren2021zerooffload}, big-batch training~\cite{you2019large} and aggressive communication quantization~\cite{dettmers20168bit}.
All of these combined reduce time and frequency of the communication rounds, which in turn makes training on heterogeneous devices and low-bandwidth networks possible.

\subsection{Distributed Spot Training}
In this paper, we focus only on models that fit into the memory of a single GPU, as we are interested in utilizing data parallelism on cheaper and more readily available hardware.
However, our insights are applicable to larger models with techniques such as ZeRO offloading~\cite{ren2021zerooffload}, more aggressive quantization~\cite{wortsman2023stable} and even model parallelism~\cite{ryabinin2023swarm}.
The current options for data parallelism are either using multiple GPUs on the same node (e.g., a DGX system with eight GPUs) or having multiple nodes with a GPU each in the same high-bandwidth network (>25~Gb/s) to minimize communication time.
The latter does not work on cheap but interruptable instances, while the former has some use in the form of Amazon Sagemaker but is limited to a single node and is typically very pricey (spot pricing for DGX-2 is \$6.30/h versus 8xT4 at \$0.72/h on GC).
However, using Hivemind, a new training scenario becomes feasible: Distributed training in a decentralized fashion on interruptable VMs with bandwidths of <1~Gb/s.
Since spot instance prices change hourly depending on the time of day and zone availability~\cite{lee2017deepspotcloud}, and can vary widely between cloud providers (cf.~\Cref{tab:cloud-pricing}), training between continents and in multiple clouds could potentially be more cost-effective than using a single, more computationally powerful node at spot prices.

With the newly added training setups from~\Cref{fig:cv-sps-trade-off} (circled), it was not previously possible to choose the best option, and having the option to combine older, more available GPUs is a net benefit for both consumers as well as cloud providers.
Our paper shows that it is possible to train on multiple clouds across multiple continents and provides guidelines on how to accomplish this cost-efficiently.
\vspace*{-2mm}
\section{Model Suitability} 
\label{sec:model-suitability}

Selecting suitable models with a big enough parallel workload is essential to ensure successful distributed spot training.
To cover a wide range of established models, we drew from MLCommons' comprehensive DL training benchmark~\cite{mattson2020mlperf}.
We used models from the CV and NLP domains and gradually increased their size and TBS to increase the parallel compute amount. 
As discussed in ~\Cref{sec:distributed-dl}, the TBS may be exclusively responsible for the success of distributed training and was chosen to cover both medium and large batches (8K, 16K and 32K).
These minibatch sizes start to become more common due to the LAMB optimizer~\cite{you2019large}, which works well enough for both smaller (512) and huge batches (64K) and should be representative of state-of-the-art workloads.
{\color{diff} For a represenatative experimental study with a minibatch size of 256 on the automatic speech recognition model (Whisper~\cite{pmlr-v202-radford23a}), please refer to \Cref{sec:appendix}.}
All experiments were run with FP16 precision, as the target T4 GPUs have a considerable improvement in FLOPs compared to FP32 (8:1).

For \textbf{CV}, we take five models from the extended ResNet family, starting with the smallest one, ResNet18~\cite{he2015deep} (RN18), ResNet50 (RN50), ResNet152 (RN152), WideResNet101\_2~\cite{zagoruyko2016wide} (WRN101) and ConvNextLarge~\cite{liu2022convnet} (CONV), which is almost 20 times larger than RN18.
The paramter count is 11.7M, 25.6M, 60.2M, 126.9M, and 197.8M, respectively.
{\color{diff}These models were popularized due to their ability to help with the vanishing gradient problem by using residual connections between layers.
Currently, they are not only used for classification, but can serve as an embedding of images by removing the classification head~\cite{9373350,elharrouss2206backbones}.}
For the dataset, we use Imagenet1K~\cite{deng2009imagenet} and train the classification task, which tries to assign one of 1000 classes to each image.

For \textbf{NLP}, we selected three models from the BERT family:\\ RoBERTaBase~\cite{liu2019roberta} (RBase), -Large (RLrg), and -XLM~\cite{conneau2020unsupervised} (RXLM).
The parameter count is 124.7M, 355.4M, and 560.1M, respectively.
We used the same configuration as the original models and trained them on masked language modeling, a common pre-training task.
{\color{diff} RoBERTa models were a replication study of BERT but with a focus on better hyperparameter tuning, leading to state-of-the-art results and proposed using much higher minibatch sizes than previously common.}
The text dataset is March '22 Wikipedia~\cite{wikidump}.

When we run our experiments in a multi-cloud environment on spot instances, we cannot plug in proprietary cloud storage or wait for the dataset to download, as the instances can be terminated anytime.
To simulate a real-world deployment with a non-public dataset, we chose an independent S3 storage provider, Backblaze (B2)~\cite{b2}.
Backblaze has replicated data centers that can better serve requests from anywhere worldwide, guaranteeing a reasonable ingress rate from every continent.
Additionally, the cost is very manageable at \$0.01/GB rate for egress and \$0.005/GB/month for storage.
A detailed analysis of the costs incurred for the experiments can be found in~\Cref{sec:multicloud-performance}.
We access the datasets on-demand via shards in the \texttt{tar} format with the WebDataset library~\cite{aizman2019webdataset}.
We chose WebDataset due to its features like automatic local caching, streaming decompression, streaming preprocessing, and having an easy to work with archive format that allows representing the data in its original format.
Finally, for the Hivemind parameterization, we enabled delayed parameter averaging (DPU)~\cite{ren2021zerooffload} to enable simultaneous gradient communication and computation at the expense of a round of staleness. We selected FP16 compression for peer-to-peer communication.

\textbf{Experimental design.} First, we must verify that our models are suitable for cloud training.
For this purpose, we evaluate them on the powerful Ampere GPUs first - if they scale there without facing a communication bottleneck, they should also scale on the slower T4, which is common at GC, AWS, and Azure.
We use the LambdaLabs~\cite{lambdaweb} for these experiments, which gives us on-demand A10 GPUs for just \$0.60/hour, but currently offer their services only in the US West region.
All experiments are performed on the 515.65.01 driver, CUDA 11.6, and PyTorch 1.13.1.
We profiled a network bandwidth of 3.3~Gb/s and a latency of 0.3 ms between the Lambda VMs. 
   
To establish a fair baseline, we train all models from ~\Cref{tab:experimental-setup-for-model-suitability} on a single GPU that achieves large minibatch sizes through gradient accumulation.
Processes logs system metrics every second and evaluates the training performance whenever a batch is processed.
Finally, all multi-GPU experiments are monitored with a training monitor that scrapes the DHT every second to log the peer state and training progress synchronously.

\begin{figure} 
    \begin{subfigure}[c]{0.22\textwidth} 
        \includegraphics[width=\textwidth]{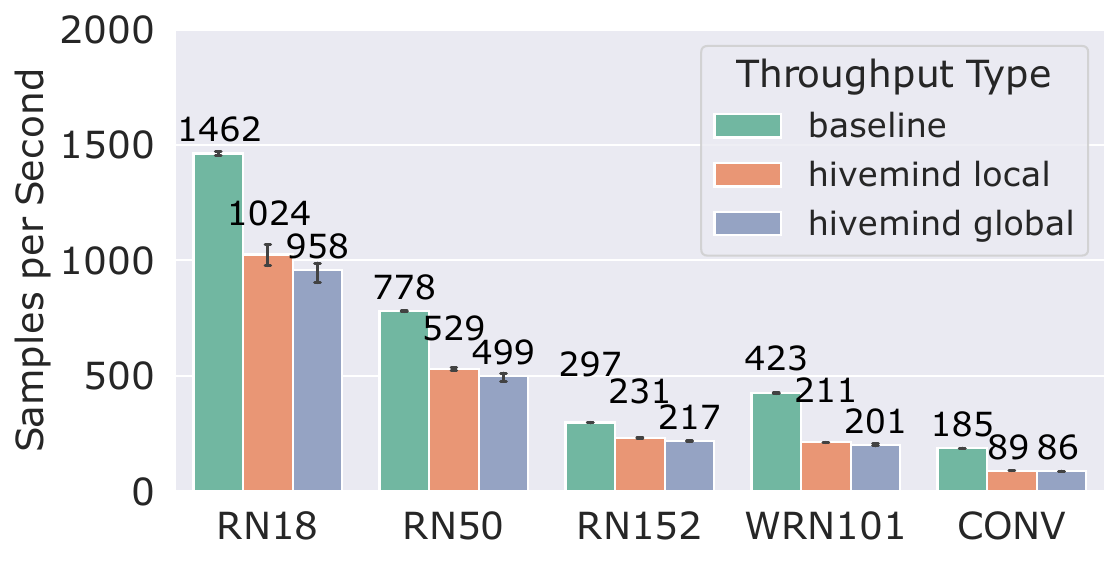}
        \vspace{-5mm}  
        \caption{CV}
        \label{fig:cv-2xa10-local-sps}
    \end{subfigure} 
    \begin{subfigure}[c]{0.23\textwidth}
        \includegraphics[width=\textwidth]{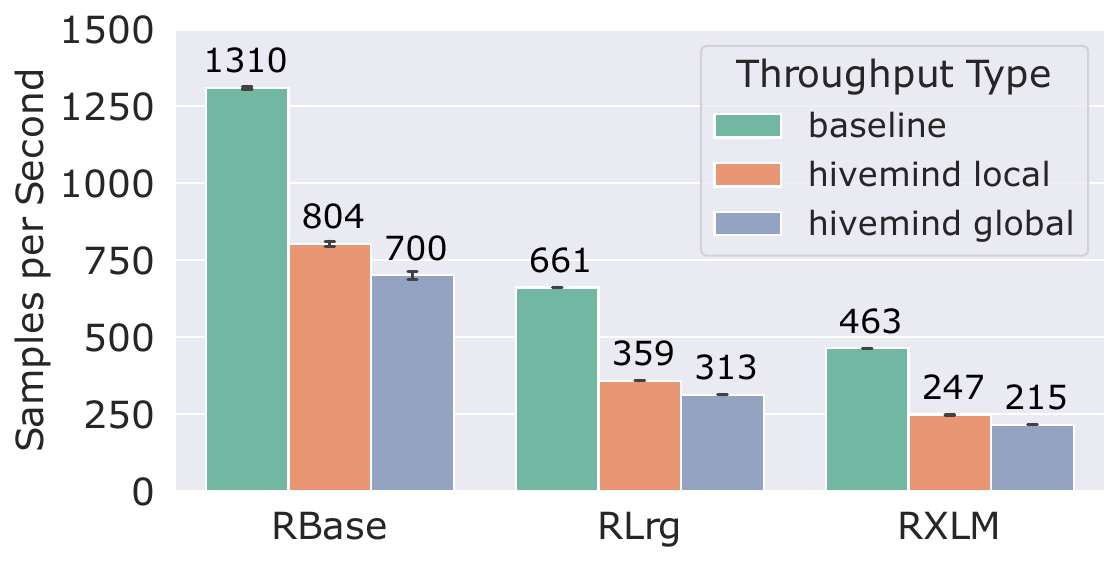}
        \vspace{-6mm}
        \caption{NLP}
        \label{fig:nlp-2xa10-local-sps}   
    \end{subfigure}
    \vspace{-4mm}
    \caption{Hivemind penalty on normalized throughputs.}
    \label{fig:hivemind-penalty-analysis}
    \vspace*{-2mm}
\end{figure}

\textbf{(1) Hivemind penalty.} Using Hivemind as middleware to share gradients and keep a fully decentralized architecture running harms performance compared to single-node training. 
We can compare the effects of Hivemind training by looking at three metrics: \textit{baseline}, the single GPU throughput, \textit{hivemind local}, normalized GPU throughput without the averaging step, and \textit{hivemind global}, the actual normalized GPU throughput.
When comparing the baseline and local speed in ~\Cref{fig:hivemind-penalty-analysis} for a setup with two GPUs, running Hivemind reaches at best 78\% (RN152) and at worst 48\% (CONV) of the baseline performance.
Unsurprisingly, the larger the model size, the worse the penalty gets due to the increased size of the accumulated gradients (GAC) over each step.
However, the baseline also applies gradient accumulation to reach the target minibatch size without the performance drop.
After isolating the respective function calls, there seems to be a slight inefficiency in how GAC is implemented in Hivemind versus the native PyTorch call.
We are working with the maintainers to fix this issue~\cite{issue566}.
On the other hand, the disadvantage of synchronization is minimal under the perfect conditions of a good interconnect.
The global speed in ~\Cref{fig:cv-2xa10-local-sps,fig:nlp-2xa10-local-sps} only degrades at best to 97\% (CONV) to at worst to 87\% (RBase) compared to the local throughput, meaning that the communication under these conditions only accounts for a fraction of the total training time.
This degradation is inversely correlated to the model size due to larger models training quadratically longer per parameter, but the communication only increases linearly~\cite{ryabinin2023swarm}.
While an implementation issue currently affects performance, and the worst total performance drop is at 47\% (CONV baseline vs. global), scaling is still possible with a ratio of roughly 2:1 of GPUs to throughput.
We further refine this ratio in the following section by comparing which models are most suitable to be trained in a distributed environment.
 
\begin{figure}
    \begin{subfigure}[c]{0.23\textwidth}
        \includegraphics[width=\textwidth]{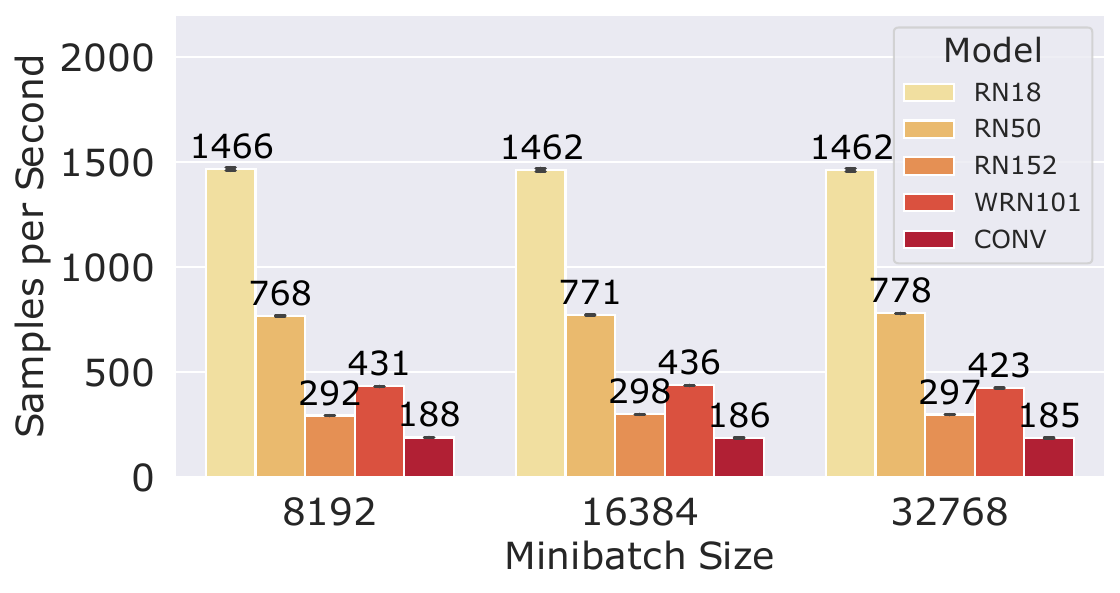}
        \vspace{-6mm}
        \caption{CV 1xA10}
        \label{fig:cv-1xa10-baseline}
    \end{subfigure}
    \begin{subfigure}[c]{0.23\textwidth}
        \includegraphics[width=\textwidth]{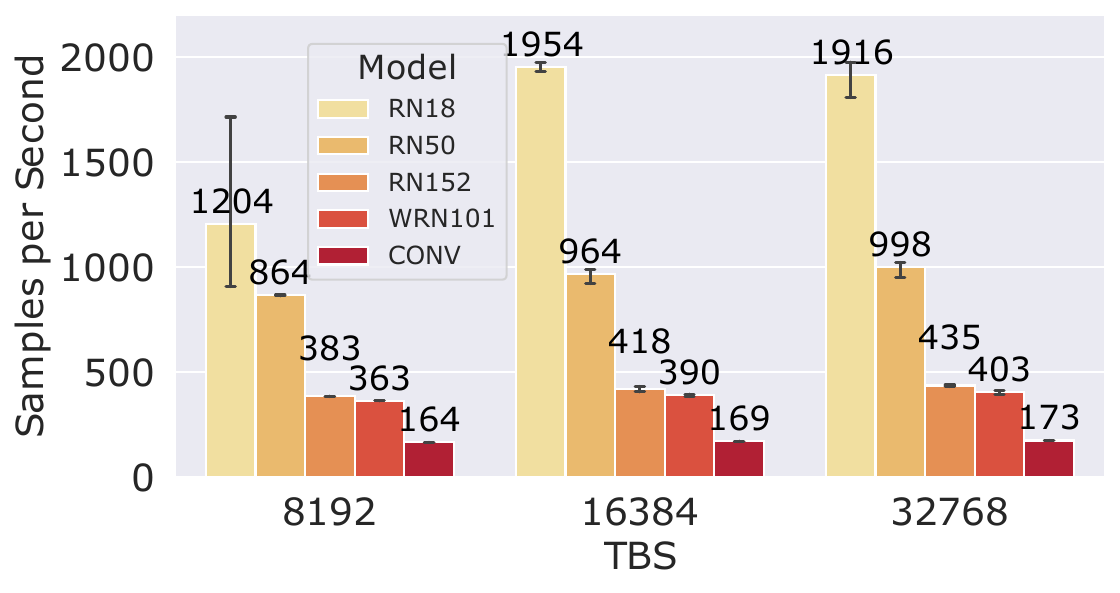}
        \vspace{-6mm}
        \caption{CV 2xA10}
        \label{fig:cv-2xa10-hivemind}
    \end{subfigure}
        \begin{subfigure}[c]{0.23\textwidth}
        \includegraphics[width=\textwidth]{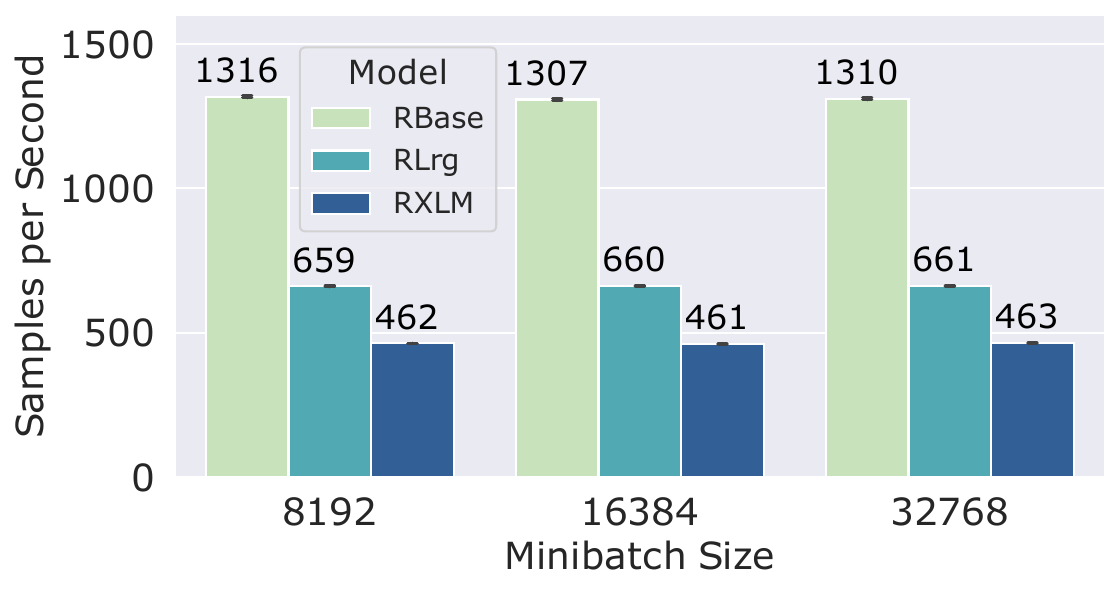}
        \vspace{-6mm}
        \caption{NLP 1xA10}
        \label{fig:nlp-1xa10-baseline}
    \end{subfigure}
    \begin{subfigure}[c]{0.23\textwidth}
        \includegraphics[width=\textwidth]{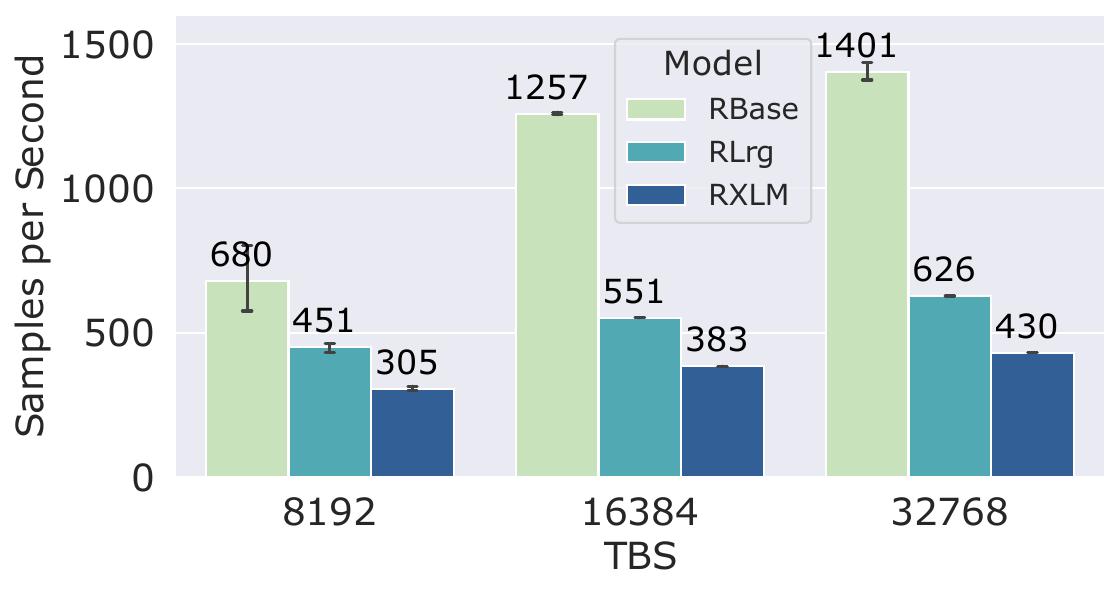}
        \vspace{-6mm}
        \caption{NLP 2xA10} 
        \label{fig:nlp-2xa10-hivemind}
    \end{subfigure}
    \vspace{-3mm}
    \caption{Throughput comparison between single GPU baselines and the Hivemind runs with two GPUs.}
    \label{fig:1x-vs-2xa10-throughput-comparison}
    \vspace*{-5mm}
\end{figure}
\textbf{(2) Less suitable models for distributed spot training.} While training billion-parameter NLP models scale well due to the "square-cube" law, the minimum model size is not yet fully defined~\cite{ryabinin2023swarm}. 
The reason is that many factors play a role in whether a model is suited for geo-distributed training.
On the one hand, a small model results in small gradients exchanged between peers, so the averaging step is fast.
On the other hand, a small model will also reach the TBS faster than larger models, which may lead to a low speedup if the calculation time is disproportionally lower than the communication time.

We found the granularity metric~\cite{10.5555/541880}, typically used in high-performance computing, practical to attach a comparable value to each setup to quantify the ratio of the calculation to communication time.
The higher the granularity, the more parallelizable the task, as more calculation can be distributed between peers, ensuring a good speedup.
It is important to note that this metric depends on the model and the hardware being used. 
The communication time is affected by the parameter count, and the calculation time is affected by the layer type of the parameters (including feedforward, convolution, and transformer).
Therefore, the calculation time can decrease with improved hardware, which we evaluate in ~\Cref{sec:hybrid-cloud-performance}.
Another parameter that affects the calculation time is the TBS that all peers work to accumulate.
There is a practical limit to the TBS where a model is still trainable, which is currently at 64K with the LAMB optimizer~\cite{you2019large}.
This limits the possibility of improving the speedup of small models by increasing the batch size, meaning that at some point, the speed will be limited by the communication time. 
It is important to remember that just increasing the TBS to create more calculation time can have a grave effect on training performance if the optimizer is not adequately selected and configured.

Our experimental results in ~\Cref{fig:1x-vs-2xa10-throughput-comparison} show the practical implications of this observation.
For the 2xGPU experiments in ~\Cref{fig:cv-2xa10-hivemind,fig:nlp-2xa10-hivemind}, we can see the effect of a TBS increase which improves the total throughput.
Doubling the TBS equals cutting down the per-sample communication cost by two, which leads to the slight increase in performance visible in both CV and NLP experiments.
However, the smallest models, RN18 and RBase, fluctuate significantly at a TBS of 8K due to a minimum matchmaking time of 5 seconds. 
Whenever all peers accumulate the TBS in less than 5 seconds, the asynchronous thread that matches the peers in groups to perform the all-reduce may still need to finish.
This results in an unstable averaging time, which limits the scalability of small models with a small TBS.

\begin{figure}
    \begin{subfigure}[c]{0.23\textwidth}
        \includegraphics[width=\textwidth]{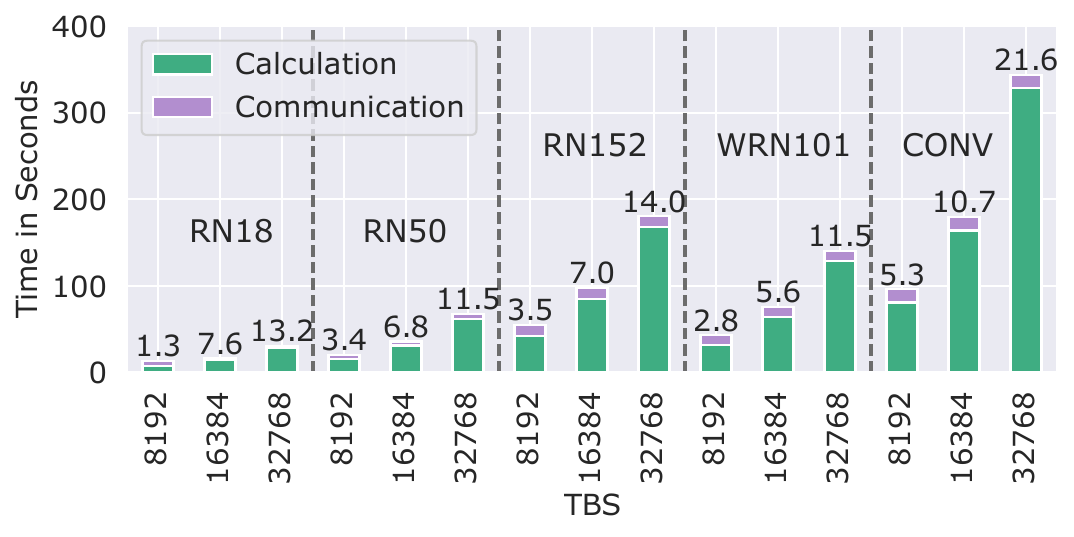}
        \vspace{-6mm}
        \caption{CV}
        \label{fig:cv-2xa10-granulartiy}
    \end{subfigure}
    \begin{subfigure}[c]{0.23\textwidth}
        \includegraphics[width=\textwidth]{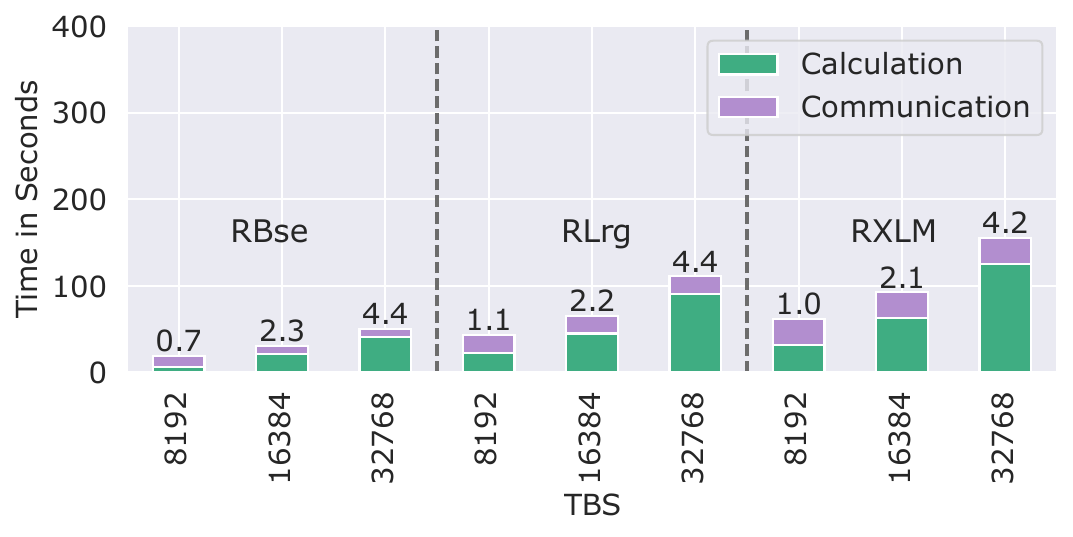}
        \vspace{-6mm}
        \caption{NLP} 
        \label{fig:nlp-2xa10-granularity}
    \end{subfigure}
    \vspace{-3mm}
    \caption{TBS vs. total training time on 2xA10s. Granularity is shown above each bar. Dotted lines separate different models.}
    \label{fig:2xa10-granulartiy}
    \vspace*{-5mm}
\end{figure}

To illustrate how the TBS and model size affect the individual timings, we visualize the total training time split up into the calculation and communication time in ~\Cref{fig:2xa10-granulartiy}.
CV models are generally computationally more expensive and have a higher granularity than NLP models, which have slightly longer averaging rounds due to the much larger model sizes. 
When comparing the models at the same TBS (e.g., 32K), there is an inconclusive relation between runtime and parameter count.
Some models increase their runtime with parameter count w.r.t. smaller models (RN50 to RN152, RBase to RLrg), while others decrease their runtime (RN152 to WRN101, RLrg to RXLM).
This performance is due to not all layer parameters contributing similarly to computational complexity.
Depending on the specific architecture, even models with more parameters can be faster to train due to a more efficient architecture, such as the WRN101~\cite{zagoruyko2016wide}.

The communication time between different TBS sizes stays the same, barring the two matchmaking time exceptions (RN18, RBase), as the gradients are accumulated before being sent.
For all other models, doubling the TBS leads to exactly double the amount of work and doubles the granularity. 
With a TBS of 32K, all models have a granularity of at least 4.2 (RXLM) and at most 21.6 (CONV), which show strong scaling potential. 
Therefore, we decided to use a TBS of 32K for all following experiments to ensure that the setup scales before introducing bandwidth and computational limitations.

Summarizing, whether a model is scalable without network bandwidth limitations depends on the minimum time to reach the TBS and on the granularity.
Tuning the TBS is possible to a certain extent but depends on the specific training task and optimizer configuration.

\begin{figure}
    \begin{subfigure}[c]{0.23\textwidth}
        \includegraphics[width=\textwidth]{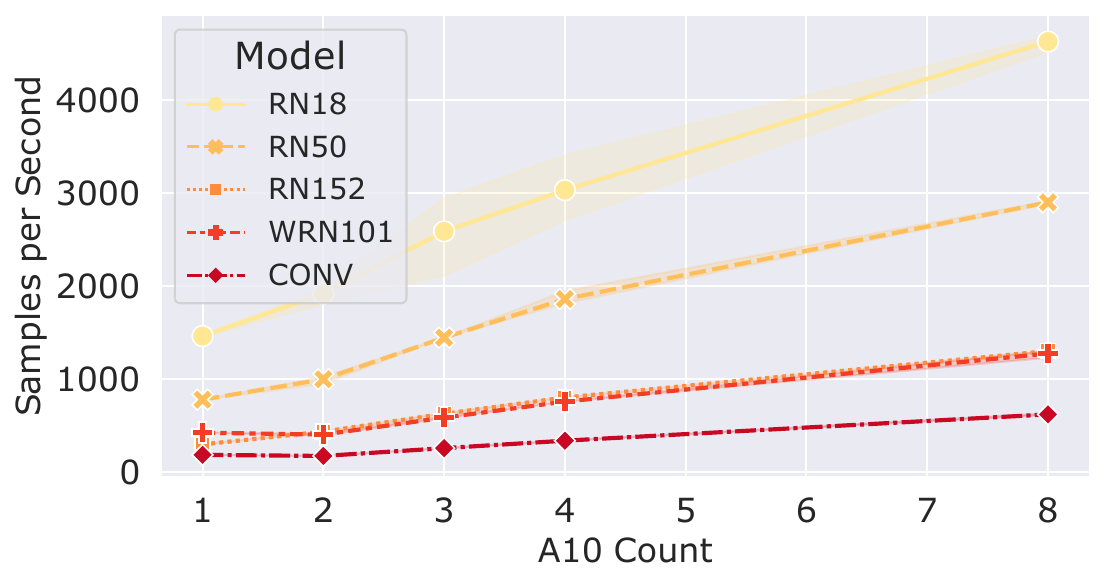}
        \vspace{-6mm}
        \caption{CV}
        \label{fig:cv-multi-a10}
    \end{subfigure}
    \begin{subfigure}[c]{0.23\textwidth}
        \includegraphics[width=\textwidth]{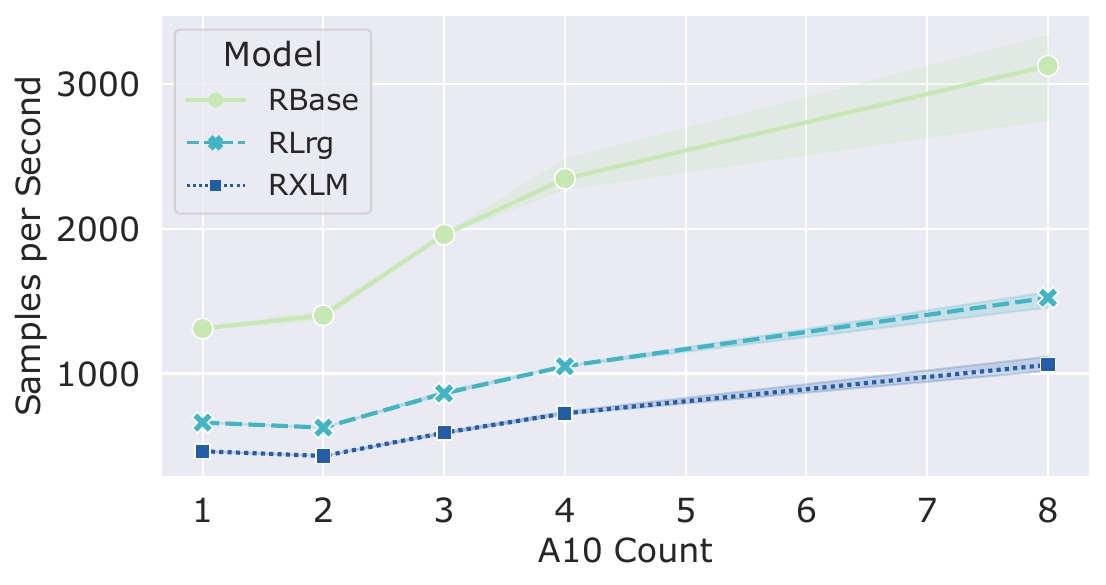}
        \vspace{-6mm}
        \caption{NLP}
        \label{fig:nlp-multi-a10}  
    \end{subfigure}
    \vspace{-3mm}
	\caption{Throughput comparison from 1 to 8 A10 GPUs.}
	\label{fig:multi-a10}
    \vspace*{-4mm}
\end{figure}

  
\begin{figure}
    \begin{subfigure}[c]{0.23\textwidth}
        \includegraphics[width=\textwidth]{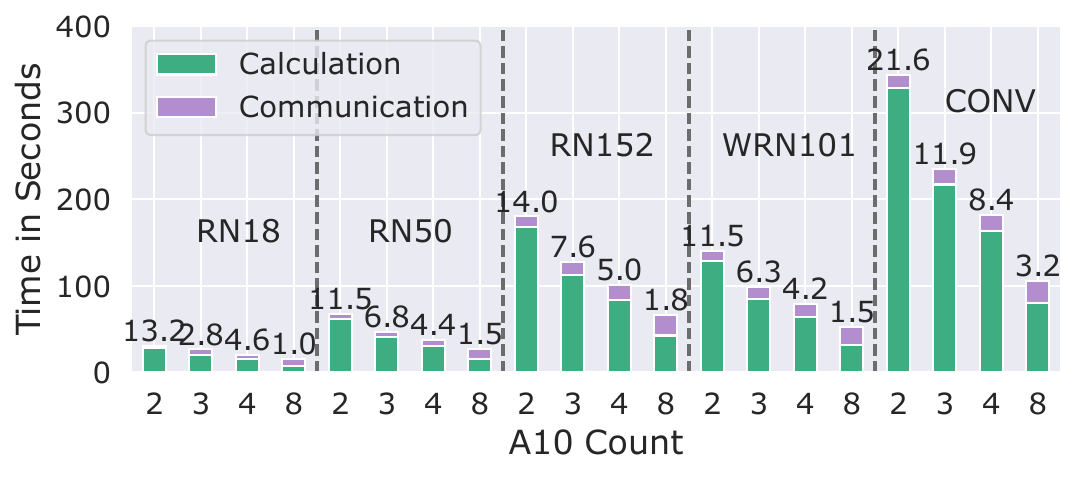}
        \vspace{-6mm}
        \caption{CV} 
        \label{fig:cv-multi-a10-granularity}
    \end{subfigure}
    \begin{subfigure}[c]{0.23\textwidth}
        \includegraphics[width=\textwidth]{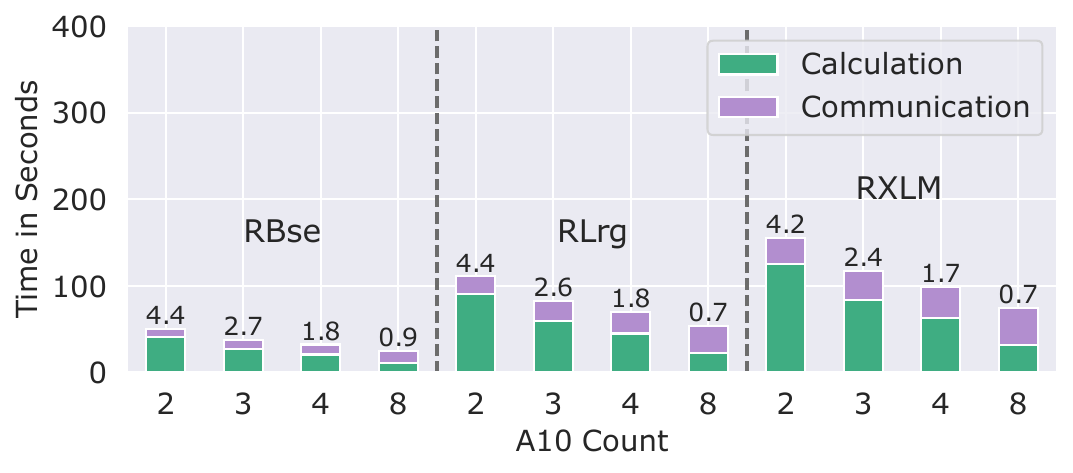}
        \vspace{-6mm}
        \caption{NLP}
        \label{fig:nlp-multi-a10-granularity}
    \end{subfigure} 
    \vspace{-3mm} 
    \caption{Multi-GPU scalability at 32K TBS. Granularity is shown above each bar. Dotted lines separate different models.}
    \label{fig:multi-a10-granulartiy}
    \vspace*{-3mm}
\end{figure}  

\textbf{(3) Per-GPU speedup decreases with low granularity.}
To evaluate the scalability with additional hardware, we profile all models on 2,3,4, and 8 GPUs with a TBS of 32K.
~\Cref{fig:multi-a10} shows the throughput for all models in the different hardware scenarios. 
Generally, all models scale well regardless of size, with the best speedup of 4.37x (RN152) and the lowest at 2.29x (RXLM) with 8 GPUs.
There is a visible trend in the per-GPU contribution to the speedup ($\frac{\text{speedup}}{\#\text{GPUs}}$).
The more GPUs we add, the lower the contribution, e.g., RN18 goes from 0.7 to 0.4 with two to eight GPUs, respectively.
This decrease is likely to continue due to a granularity of 1.0 at 8 GPUs (\Cref{fig:cv-multi-a10-granularity}), as doubling the GPUs would, at best, increase the throughput by 33\% by halving the calculation time.
However, the more computationally expensive the models are, the slower the per-GPU contribution falls off and the larger the granularity is (RN152, CONV).
This does not hold true for our NLP models (\Cref{fig:nlp-multi-a10-granularity}); while they have increasingly more model parameters, the only difference between the two biggest models, RLrg and RXLM, is the vocabulary size increase of 50K to 250K.
Due to how embedding layers are lookups, the forward pass is not affected by the increased embedding size, but the backward pass is.
This results in a smaller increase of the calculation time while communication increases linearly with the number of parameters.

Additionally, we see the drop in throughput when comparing the single GPU and dual GPU experiments for most larger models (\Cref{fig:multi-a10}), which stems from observation \textbf{(1)} of the Hivemind penalty.  

We also observe that with each subsequent doubling of GPUs, the calculation time is halved, while the communication increases sub-linearly due to the more efficient group-based all-reduce of MoshpitSGD~\cite{ryabinin2021moshpit}.
For example, the averaging step for the RXLM on 2xA10 takes 5 seconds per GPU (10s total), while the 8xA10 averaging step takes 1.8 seconds per GPU (14.4s total).

In summary, all models show a speedup but have a decreasing per-GPU contribution due to smaller granularity with more GPUs.
Therefore, the larger the model and TBS, the greater the scaling potential.
High granularity is a good indicator of scalability, and since the communication time only increases linearly with additional peers (cf.~\Cref{ssec:hivemind}), knowing the initial calculation time is a good indicator of future throughput.
Under the optimal conditions of good compute performance and an interconnect with relatively high bandwidth, scaling was not a problem.
But what happens under less favorable conditions in geo-distributed settings?

\section{Geo-distributed Performance}
\label{sec:geodistributed-performance} 
\begin{table}[]
    \caption{Geo-distributed experiments on GC with T4 VMs.}
    \vspace*{-2mm}
    \scalebox{0.7}{
    \begin{tabular}{r|l|r}
    \textbf{Exp. Name} & \textbf{Resources} & \textbf{Total} \\ \hline
    \textbf{A-\{1,2,3,4,6,8\}} & \{1, 2, 3, 4, 6, 8\}xUS     & 1,2,3,4,6,8\\ \hline
    \textbf{B-\{2,4,6,8\}} & \{1, 2, 3, 4\}xUS + \{1, 2, 3, 4\}xEU & 2,4,6,8\\ \hline
    \textbf{C-\{3,6\}} & \{1, 2\}xUS + \{1, 2\}xEU + \{1, 2\}xASIA & 3,6\\
    \textbf{C-\{4,8\}} & \{1, 2\}xUS + \{1, 2\}xEU + \{1, 2\}xASIA + \{1, 2\}xAUS & 4,8\\
    \end{tabular}
    }
    \label{tab:geodistributed-experiments}
    \vspace*{-7mm}
\end{table} 

As spot prices for the same hardware differ depending on the region, zone, and time of day~\cite{lee2017deepspotcloud}, it might be a good idea to use VMs across different data centers.
However, is the connectivity between regions and continents good enough to enable distributed deep learning?
To explore this question, we decided to conduct three types of experiments (\Cref{tab:geodistributed-experiments}):
\begin{description}
    \item[(A) Intra-zone] Can we scale if the VMs are co-located in the same zone (\texttt{us-central-1})?
    \item[(B) Transatlantic] Can we scale when we combine VMs from two regions (US and EU), and what happens when the compute is unevenly distributed across regions?
    \item[(C) Intercontinental] Can we scale if we combine VMs from four continents (US, EU, ASIA, AUS)?
\end{description}

\textbf{Experimental design.} 
Based on the insights from ~\Cref{sec:model-suitability}, we decided to use the largest models (CONV, RXLM) for all further cloud experiments in ~\Cref{sec:geodistributed-performance,sec:multicloud-performance,sec:hybrid-cloud-performance}, with the TBS of 32K as a baseline with good scaling properties.
We abbreviate them with their respective domain names (CV, NLP).
We used Google Cloud~\cite{gcweb} for all experiments in this section, as they were the first to give us access to all necessary zones.
The default networking solution in GC is the "Premium Tier", which tries to use a Google-owned network instead of the public internet.
We measured the throughput and latency between all zones via \texttt{iperf} and \texttt{ping} and report the average of 5 consecutive runs in ~\Cref{tab:geodistributed-network}.
Unsurprisingly, the diagonal shows that the local connectivity between zones runs at almost 7~Gb/s with a latency of 0.7ms, probably due to the hypervisors being in the same data center.
While the up- and download were perfectly symmetrical in all setups, the throughput dropped to <210~Mb/s for all non-local connections.
The US-based data center is located in Iowa and is best connected with at least 120~Mb/s to the remaining regions, namely Belgium in the EU (6,911km), Taiwan in ASIA (11,853km), and Sydney in Australia (AUS, 14,555km), presumably due to the physical distance. 
The lowest bandwidth and highest latency connections are between the EU region and ASIA and AUS, reaching around 80~Mb/s and 270ms.
We decided to use the \texttt{n1-standard-8} template with eight cores, 30~GB memory, and a T4 GPU, as the smaller image with 15~GB was insufficient to meet the memory requirements for gradient application on the CPU with the biggest models.
The experiment naming in this section is prefixed with the type of location \textbf{(A)}, \textbf{(B)} or \textbf{(C)} and the number of VMs, e.g., A-4 is the intra-zone experiment with 4 VMs.
The full experimental description is specified in ~\Cref{tab:geodistributed-experiments}.
\begin{table}[h]
    \caption{Throughput and latency between GC zones.}
    \vspace*{-3mm}
    \scalebox{0.6}{
    \begin{subtable}[h]{0.38\textwidth}
        \centering
        \caption{Single stream TCP throughput in Gb/s.}
        \begin{tabular}{l|rrrr}
        \backslashbox{\textbf{From}}{\textbf{To}} & \textbf{US} & \textbf{EU} & \textbf{ASIA} & \textbf{AUS} \\ \hline
        \textbf{US}   & \cellcolor[HTML]{67F13A}6.90 & \cellcolor[HTML]{fadc24}0.21 & \cellcolor[HTML]{fab724}0.13 & \cellcolor[HTML]{fab724}0.12 \\
        \textbf{EU}   & \cellcolor[HTML]{fadc24}0.21 & \cellcolor[HTML]{67F13A}6.81 & \cellcolor[HTML]{fa7e24}0.08 & \cellcolor[HTML]{fa7e24}0.07 \\
        \textbf{ASIA} & \cellcolor[HTML]{fab724}0.13 & \cellcolor[HTML]{fa7e24}0.08 & \cellcolor[HTML]{67F13A}6.79 & \cellcolor[HTML]{fab724}0.16 \\
        \textbf{AUS}  & \cellcolor[HTML]{fab724}0.12 & \cellcolor[HTML]{fa7e24}0.07 & \cellcolor[HTML]{fab724}0.16 & \cellcolor[HTML]{67F13A}6.84 
        \end{tabular}
        \vspace{1mm}
        \label{tab:geodistributed-throughput}
    \end{subtable}
    }
    \scalebox{0.6}{
    \begin{subtable}[h]{0.38\textwidth}
        \centering
        \caption{ICMP latency in ms.}
        \begin{tabular}{l|rrrr}
        \backslashbox{\textbf{From}}{\textbf{To}}     & \textbf{US} & \textbf{EU} & \textbf{ASIA} & \textbf{AUS} \\ \hline
        \textbf{US}   & \cellcolor[HTML]{67F13A}0.66 & \cellcolor[HTML]{fadc24}103.11 & \cellcolor[HTML]{fab724}157.09 & \cellcolor[HTML]{fab724}176.19 \\
        \textbf{EU}   & \cellcolor[HTML]{fadc24}103.14 & \cellcolor[HTML]{67F13A}0.65 & \cellcolor[HTML]{fa7e24}253.10 & \cellcolor[HTML]{fa7e24}271.98 \\
        \textbf{ASIA} & \cellcolor[HTML]{fab724}157.08 & \cellcolor[HTML]{fa7e24}253.09 & \cellcolor[HTML]{67F13A}0.72 & \cellcolor[HTML]{fab724}131.45 \\
        \textbf{AUS}  & \cellcolor[HTML]{fab724}175.98 & \cellcolor[HTML]{fa7e24}272.08 & \cellcolor[HTML]{fab724}131.42 & \cellcolor[HTML]{67F13A}0.64
        \end{tabular}
        \vspace{1mm}
        \label{tab:geodistributed-ping}
    \end{subtable}
    }
    \label{tab:geodistributed-network}
    \vspace*{-3mm}
\end{table}

\begin{figure}
    \begin{subfigure}[c]{0.25\textwidth}
        \includegraphics[width=\textwidth]{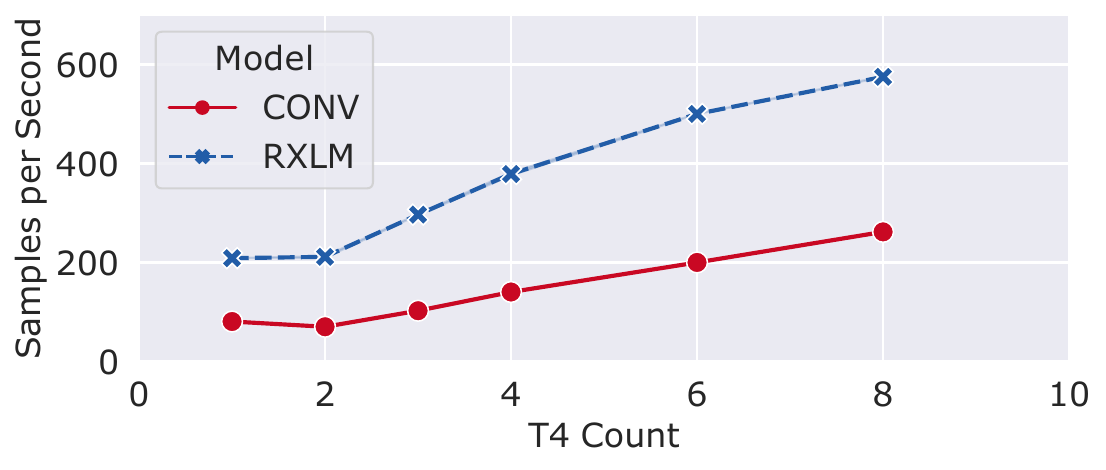}
        \vspace{-6mm}
        \caption{Throughput}
        \label{fig:geo-dist-us-only-throughput}
    \end{subfigure}
    \begin{subfigure}[c]{0.22\textwidth}
        \includegraphics[width=\textwidth]{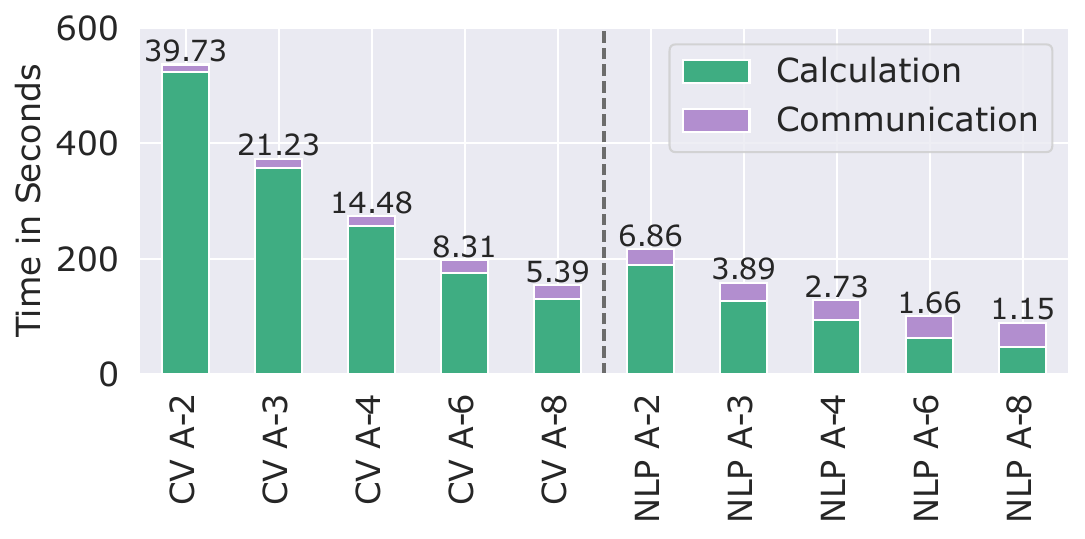}  
        \vspace{-6mm}
        \caption{Granularity}
        \label{fig:geo-dist-us-only-granularity}
    \end{subfigure}
    \vspace{-3mm}
    \caption{(A) Intra-zone performance for CV and NLP.}
    \label{fig:geo-dist-us-only}
    \vspace*{-2mm}
\end{figure}

\textbf{(A) Intra-zone scalability.}
Figure~\ref{fig:geo-dist-us-only} shows the result of the intra-zone experiments, which we used as a baseline to compare geo-distributed deployments to.
As the scalability of the CV and NLP models was already shown with much better hardware and slightly worse network connectivity (cf.~\Cref{sec:model-suitability}), the scalability with the T4 GPUs is not too surprising.
We do not see an improvement in throughput for two GPUs for either model due to the Hivemind penalty discussed in ~\Cref{sec:model-suitability}.
However, starting with three GPUs, we see an increase in throughput with a maximum speedup of up to 3.2x CV and 2.75x for NLP at eight GPUs. 
CV's per-GPU speedup ($\frac{\text{speedup}}{\#\text{GPUs}}$) is almost linear (0.43, 0.42, 0.43, 0.41, 0.41), while NLP starts dropping off faster (0.51, 0.47, 0.45, 0.40, 0.34) for 2, 3, 4, 6 and 8 GPUs, respectively.
The reason for this is the NLP granularity of 1.15 with 8 GPUs indicating an almost equal part in communication and calculation (\Cref{fig:geo-dist-us-only-granularity}) due to the much longer averaging round related to the model size (198M vs. 560M parameters).
\begin{figure}
    \begin{subfigure}[c]{0.25\textwidth}
        \includegraphics[width=\textwidth]{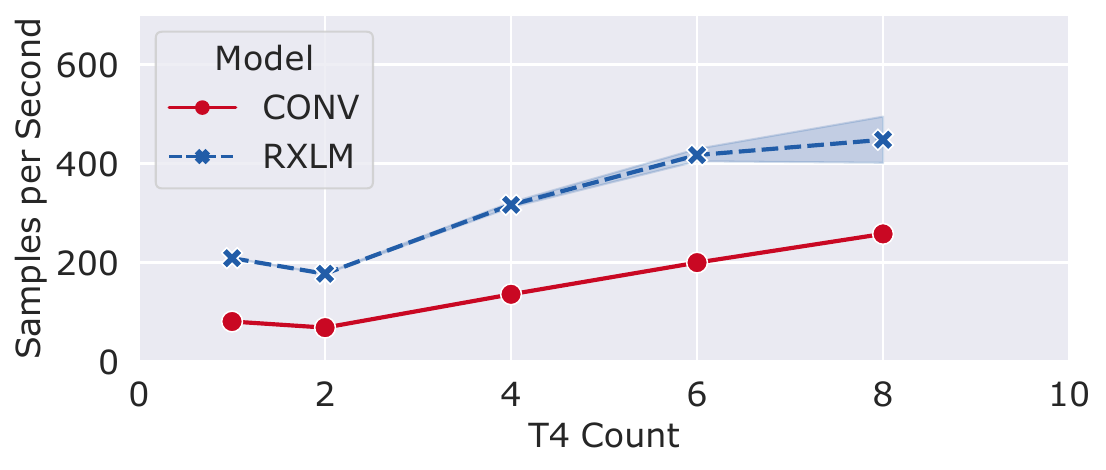}
        \vspace{-6mm}
        \caption{Throughput}
        \label{fig:geo-dist-us-eu-throughput}
    \end{subfigure}
    \begin{subfigure}[c]{0.22\textwidth}
        \includegraphics[width=\textwidth]{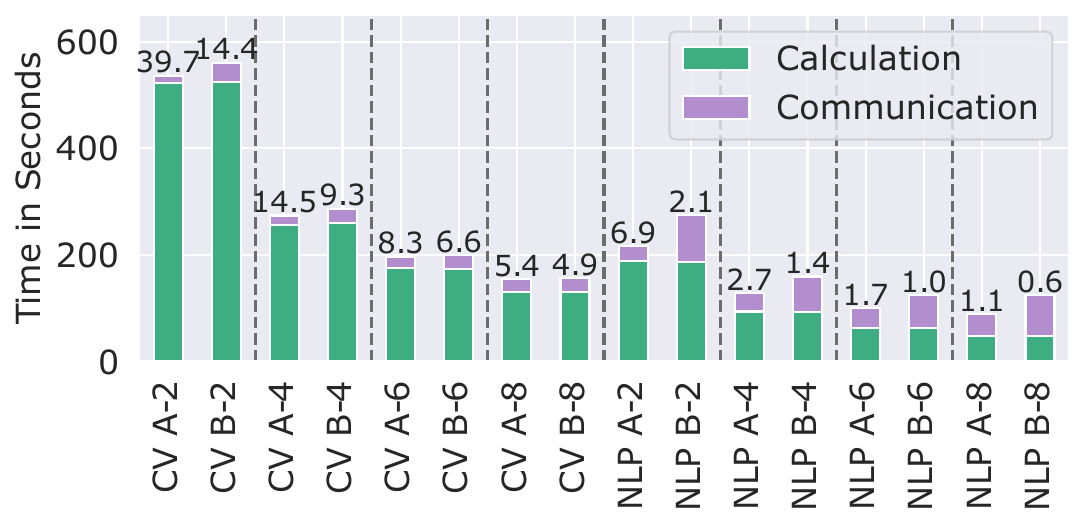}  
        \vspace{-6mm}
        \caption{Granularity}
        \label{fig:geo-dist-us-eu-granularity}
    \end{subfigure}
    \vspace{-3mm}
    \caption{(B) Transatlantic performance for CV and NLP.}
    \label{fig:geo-dist-us-eu}
    \vspace*{-2mm}
\end{figure}
The peak network bandwidth utilization between peers was at most a symmetric 1.1~Gb/s while averaging and 33~Mb/s ingress while training due to data loading.
This means that the network bandwidth of 7~Gb/s was not a limiting factor.

\textbf{(B) Transatlantic scalability.} We scale when computing hardware is local. However, what happens when there is cheap capacity in another region? In this case, we study the throughput of experiments with resources in the \texttt{us-west} and \texttt{eu-central} regions (B-2,4,6,8).

The B-2 experiment has one VM in the US and one in the EU, achieving a virtually identical throughput of 68.4 (US-EU) versus 70.1 (US) at CV (\Cref{fig:geo-dist-us-eu-throughput}). 
Our maximum peak egress rate of 250~Mb/s does not affect the CV experiments, while the US experiments peaked at 1.1 Gb/s.
The reduction in bandwidth penalizes NLP harder, where we are 16\% slower with 177.3~SPS (US-EU) compared to the intra-zone experiment with 211.4~SPS (US).
The resulting increased communication can be easily seen in the granularity analysis in ~\Cref{fig:geo-dist-us-eu-granularity} (NLP A-2,4,6,8 vs. B-2,4,6,8).
As only communication time increases in the NLP \textbf{(B)} experiments compared to \textbf{(A)}, a granularity of $\gg1$ indicates good scalability: 
Adding two more GPUs to the B-6 experiment with a granularity of 1.03 results in a throughput increase of 15\% (B-8) relative to the baseline.
Meanwhile, adding two more GPUs to the B-2 experiment with a granularity of 2.21 results in a throughput increase of 77\% (B-4) relative to the baseline.

In the B-4 experiment, we look at what happens when we increase the number of VMs to four, with two in the US and two in the EU.
Nothing surprising happens with CV, as the workload continues to be mostly computation, with a throughput of 135.8 (B-4), only 3\% slower than the intra-zone experiment with 140.4 SPS (A-4).
However, at NLP, things get more interesting as we now have more overall communication with four peers, but they can average locally first and only later transmit across the Atlantic.
However, compared to their A-counterparts, we do not see a difference in relative scalability with either B-4, B-6, or B-8. 
This means that training across regions \textbf{(B)} is slower, but the contribution per GPU decreases at the same rate as in training within a zone \textbf{(A)}.
The per-GPU speedup with additional hardware reduces at the same rate for either setup (between 0.05 and 0.06).
This results in two observations: First, communication overhead scales linearly with the number of peers.
Second, we only have to pay the penalty for transatlantic training once.
However, we cannot expect a significant improvement in communication efficiency when we increase the amount of available local resources.

Summarizing, with an transatlantic setup, CV achieves a virtually identical maximum speedup of 3.2x with 8 GPUs compared to A-1 (B-8 is 2\% slower than A-8), while NLP is more affected by lower network bandwidth and only achieves a speedup of 2.15x (B-8 is 22\% slower than A-8).
The transatlantic training penalty is applied once;  however, it does not affect the relative scaling with additional compute resources.
  
\textbf{(C) Intercontinental scalability. } To take geo-distribution to the extreme, we spawn VMs in up to 4 regions: USA, EU, ASIA, and AUS, to see how much worse bandwidth affects the training throughput (C-3,4,6,8 in ~\Cref{tab:geodistributed-experiments}).

\begin{figure}
    \begin{subfigure}[c]{0.25\textwidth}
        \includegraphics[width=\textwidth]{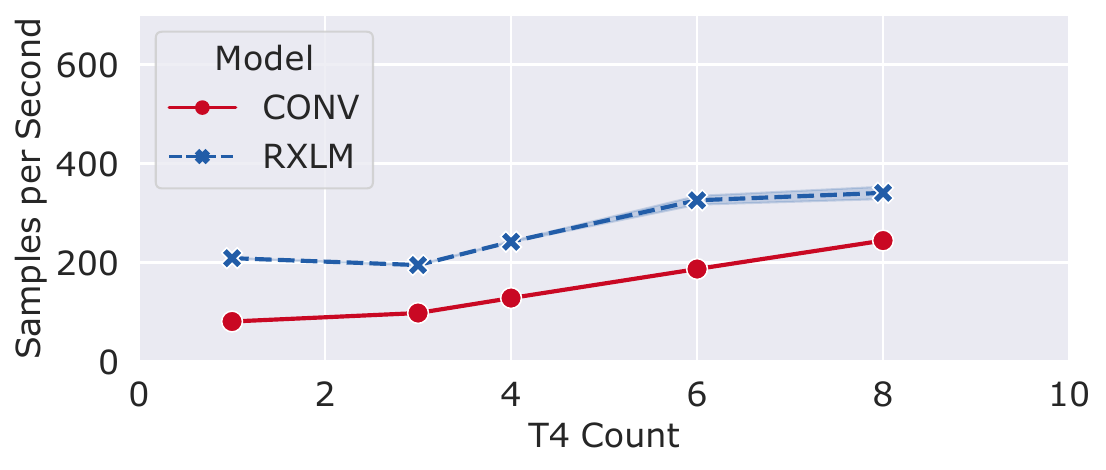}
        \vspace{-6mm}
        \caption{Throughput}
        \label{fig:geo-dist-us-eu-asia-aus-throughput}
    \end{subfigure}
    \begin{subfigure}[c]{0.22\textwidth}
        \includegraphics[width=\textwidth]{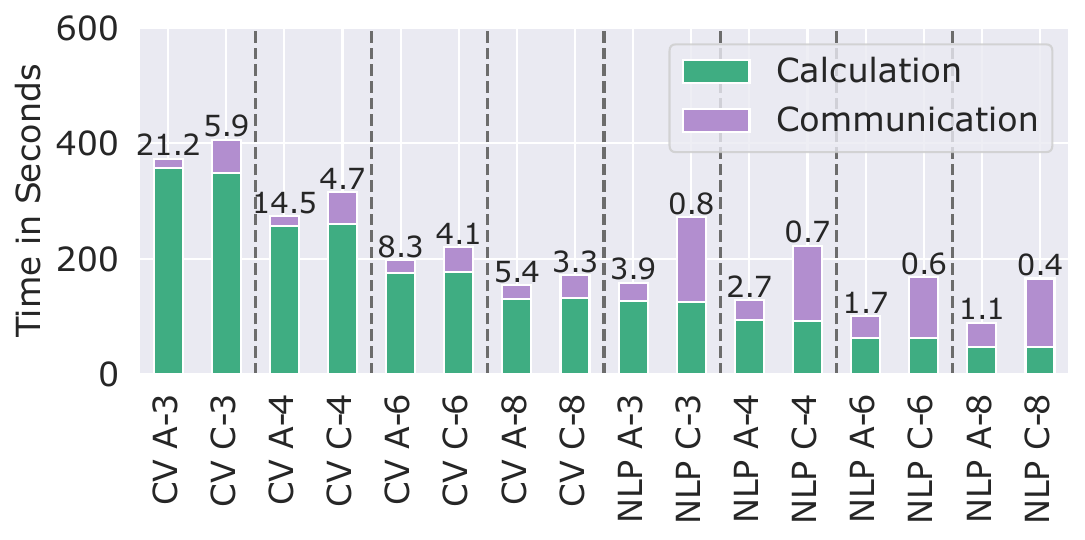}  
        \vspace{-6mm}
        \caption{Granularity}
        \label{fig:geo-dist-us-eu-asia-aus-granularity}
    \end{subfigure}
    \vspace{-3mm}
    \caption{(C) Intercontinental performance for CV and NLP.}
    \label{fig:geo-dist-us-eu-asia-oce}
    \vspace*{-6mm}
\end{figure} 

How does the intercontinental penalty investigated in \textbf{(B)} affect deployments with a single GPU on each continent?
Comparing the A-3 and C-3 experiments with three local versus three fully remote GPUs, CV is only 5\% slower, while NLP suffers a 34\% drop in throughput (\Cref{fig:geo-dist-us-eu-asia-aus-throughput}) and does not even reach the baseline single GPU performance (A-1).
The peak egress for each region was 318, 258, and 237~Mb/s for the US, EU, and ASIA, respectively.
Since our bandwidth measurements were 210 and 130~Mb/s from the US to the EU and ASIA, respectively (\Cref{tab:geodistributed-network}), this suggests that the averaging was done over the US node and not an N-to-N all-reduce (a detailed analysis of how averaging affects bandwidths is discussed in ~\Cref{sec:hybrid-cloud-performance}).
Thus, the limiting factor was the US-ASIA connection at 130~Mb/s rather than the 80~Mb/s from EU-ASIA.
The same trend continues with the C-4 run, which adds AUS as a continent with one additional VM.
As we know from the transatlantic experiments \textbf{(B)} that an additional continent has a detrimental effect on throughput, which, for the four continents experiment, C-4, results in a 9\% slower throughput for CV and 36\% slower for NLP compared to the A-4 runs (\Cref{fig:geo-dist-us-only-throughput}).
Again, the US VM is used as an averaging intermediary with a peak egress of 365~Mb/s, while the other continents are between 318 and 330~Mb/s.
When comparing the two continents (B-4) versus four continents (C-4) experiments, one GPU on each continent (C-4) is slower by 6\% for CV and 20\% for NLP compared to two GPUs on two continents (B-4).
This reinforces that local hardware should be preferred whenever possible.
However, we are always faster than the baseline (A-1), starting from 4 GPUs in both the transatlantic and intercontinental settings.
While these experiments were specifically designed to be a worst-case scenario, what about a more balanced GPU distribution with at least two GPUs in each region?

When comparing the C-6 experiment with two GPUs in three continents to the local A-6 experiments, the throughput slowdown is almost identical (CV 7\%, NLP 35\%) as with C-4 (CV 9\%, NLP 36\%) to A-4.
Scaling further to two GPUs in four continents, C-8 is slightly slower at NLP (41\%) compared to C-4 (36\%) to their respective local runs (A-8 and A-4), due to the decreasing granularity of 0.4 (\Cref{fig:geo-dist-us-eu-asia-aus-granularity}).
The small granularity removes the additional gain of four more GPUs since the task is no longer suitable for distributed training.
However, as the CV task is still at a granularity of 3.33 on C-8, it reaches a speedup of 3.02x, only 7\% slower than the fully local A-8 experiment.
The peak egress of 678~Mb/s was also reached on one US VM, while the remaining VMs were between 450 and 550~Mb/s.
These observations show that adding another continent does not significantly reduce throughput when training on three continents with at least two VMs.

In summary, while local compute is the best choice for maximum throughput, for high granularity tasks like CV, even distributing VMs over four continents only slows down performance by 7\%.
However, intercontinental training leads to a significant penalty on a task with lower granularity, like NLP, resulting in a performance drop of 41\% (C-8) compared to the fully local experiment (A-8).
Finally, each additional region introduces a constant penalty that is not amortized by adding local hardware, which should be considered when running geo-distributed training setups.

\section{Multi-cloud Performance}
\label{sec:multicloud-performance}
\begin{figure}
    \begin{subfigure}[c]{0.25\textwidth}
        \includegraphics[width=\textwidth]{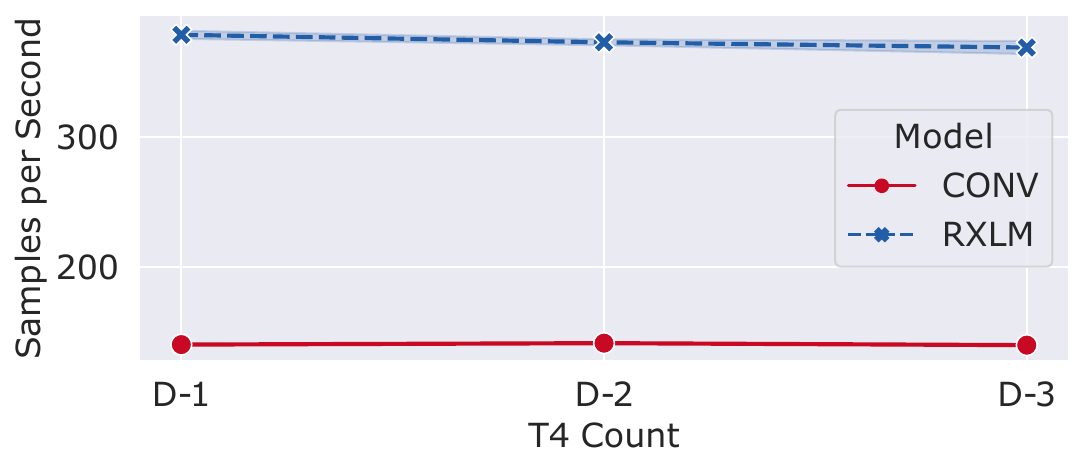}
        \vspace{-6mm}
        \caption{Throughput}
        \label{fig:multi-cloud-throughput}
    \end{subfigure}
    \begin{subfigure}[c]{0.22\textwidth}
        \includegraphics[width=\textwidth]{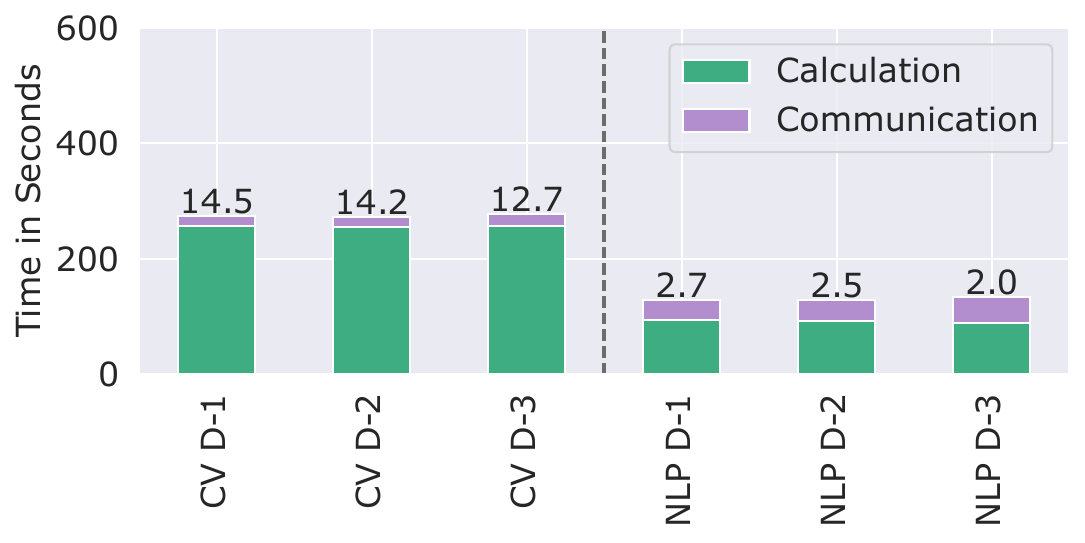}  
        \vspace{-6mm}
        \caption{Granularity}
        \label{fig:multi-cloud-granularity}
    \end{subfigure}
    \vspace{-3mm}
    \caption{Multi-cloud performance for CV and NLP.}
    \label{fig:multi-cloud-performance}
    \vspace*{-3mm}
\end{figure}
Using multiple cloud providers makes sense if we want to use resources cost-effectively and have additional reliability.
In our scenario, we are interested in what throughput per \$ can be expected and if any barriers prevent multi-cloud training.
However, one can also consider the data center's carbon footprint, which can change depending on the season and time of day~\cite{regionpicker}.
\vspace*{-2mm}
\begin{table}[h]
    \caption{Average multi-cloud throughput and latency.}
    \vspace*{-3mm}
    \scalebox{0.75}{
    \begin{subtable}[h]{0.30\textwidth}
        \centering
        \caption{Single stream TCP throughput in Gb/s.}
        \begin{tabular}{l|rrr}
        \backslashbox{\textbf{From}}{\textbf{To}}     & \textbf{GC} & \textbf{AWS} & \textbf{Azure} \\ \hline
        \textbf{GC}    & \cellcolor[HTML]{67F13A}6.35 & \cellcolor[HTML]{b2f13a}1.52 & \cellcolor[HTML]{fadc24}0.45 \\
        \textbf{AWS}   & \cellcolor[HTML]{b2f13a}1.81 & \cellcolor[HTML]{67F13A}4.87 & \\
        \textbf{Azure} & \cellcolor[HTML]{fadc24}0.47 & & \cellcolor[HTML]{67F13A}7.63
        \end{tabular}
        \vspace{1mm}
        \label{tab:multicloud-bandwidth}
    \end{subtable}
    }
    \scalebox{0.75}{
    \begin{subtable}[h]{0.30\textwidth}
        \centering
        \caption{ICMP Latency in ms.} 
        \begin{tabular}{l|rrr}
        \backslashbox{\textbf{From}}{\textbf{To}}     & \textbf{GC} & \textbf{AWS} & \textbf{Azure} \\ \hline
        \textbf{GC}    & \cellcolor[HTML]{67F13A}0.71  & \cellcolor[HTML]{b2f13a}15.3 & \cellcolor[HTML]{fadc24}51.22 \\
        \textbf{AWS}   & \cellcolor[HTML]{b2f13a}13.85 & \cellcolor[HTML]{67F13A}0.15 & \\
        \textbf{Azure} & \cellcolor[HTML]{fadc24}49.80 & & \cellcolor[HTML]{67F13A}1.56
        \end{tabular}
        \vspace{1mm}
        \label{tab:multicloud-ping}
    \end{subtable}
    }
    \label{table:multicloud-network-profile}
    \vspace*{-3mm}
\end{table}

We have compiled the current prices for spot and on-demand instances for T4 GPUs with 8 CPU cores and the egress costs for three well-known cloud providers, GC~\cite{gcweb}, AWS~\cite{awsweb}, and Azure~\cite{azureweb} (\Cref{tab:cloud-pricing}).
There are two different pricing concepts. 
On the one hand, there are GC and Azure, which offer relatively cheap instances, with 69\% and 73\% savings over on-demand pricing, respectively, and relatively expensive egress charges between continents of up to \$0.15/GB.
On the other hand, there is AWS, where the spot instance is only 51\% cheaper than the on-demand instance and more than twice as expensive as GC or Azure. 
However, the egress fees here are much cheaper at only \$0.02/GB.
Because of the additional offerings around compute, such as networking, identity and cost management, and tooling, it is not easy to fairly compare cloud providers. Therefore, we will limit ourselves to network and VM costs.

With the multi-cloud experiments from this section, we want to evaluate the following scenarios:
First, partially switching from one provider to another without stopping the training.
Second, scaling resources in the same region when one of the cloud providers is already at capacity for spot-priced VMs or the current price is too high~\cite{9975369}. We know from ~\Cref{sec:geodistributed-performance} that scaling resources in the same location can significantly improve performance, which may only be possible using additional cloud providers. 

\textbf{Experimental design}. To enable a fair comparison between the cloud providers, we rented hardware most similar to each other in the same region.
We used each provider's default settings and only changed hardware specs.
For GC, it is the same instance as in ~\Cref{sec:geodistributed-performance}. At AWS, it is a \texttt{g4dn.2xlarge} with eight cores and 32 GB in the \texttt{us-west-2c} region.
Unfortunately, we had to make two compromises with Azure.
There was only the combination of four cores and 30 GB RAM (\texttt{NC4as\_T4\_v3}), and there were no T4 GPU resources available in the \texttt{us-west}, so we had to fall back to \texttt{us-south-2}.
 
The network profiling between all cloud providers in ~\Cref{table:multicloud-network-profile} shows that their intra-cloud connectivity is comparably fast with 6.4, 4.9, and 7.6~Gb/s for GC, AWS, and Azure, respectively.
All connections are mostly symmetric, with inter-cloud connectivity between GC and AWS providing up to 1.8~Gb/s and a ping of 15.3ms, indicating that while they are likely not in the same data center, they are close to each other and connected to the same Internet exchange point.
However, connectivity to Azure could be better since it operates in a different zone, with a bandwidth of 0.5~Gb/s and a ping of 51ms.

Our experimental setup consists of four GPUs with equal contributions from each cloud provider. D-1 is the baseline with four GPUs at GC, D-2 with two GPUs each at GC and AWS, and D-3 with two GPUs at GC and Azure.
We compare moving two VMs to a different cloud provider to see the impact on cost and throughput.

\textbf{(1) No inter-cloud throughput penalty}. ~\Cref{fig:multi-cloud-performance} shows the throughput and granularity of each multi-cloud experiment.
CV and NLP runs have essentially identical throughput regardless of the combination of cloud providers.
Only the D-3 experiments show a very slight slowdown in communication time, reflected in the lower granularity score (\Cref{fig:multi-cloud-granularity}) of 12.72 in CV and 1.99 in NLP compared to the D-1 baseline scores of 14.48 and 2.73, respectively.
Actual throughput was between 1-2\% slower than the baseline, which is negligible and only related to the slightly worse connection to the Azure data center.
These results confirm our observation from ~\Cref{sec:geodistributed-performance} that network connectivity determines scalability, and one can easily train in a multi-cloud scenario.
\begin{figure}
    \begin{subfigure}[c]{0.47\textwidth}
        \includegraphics[width=\textwidth]{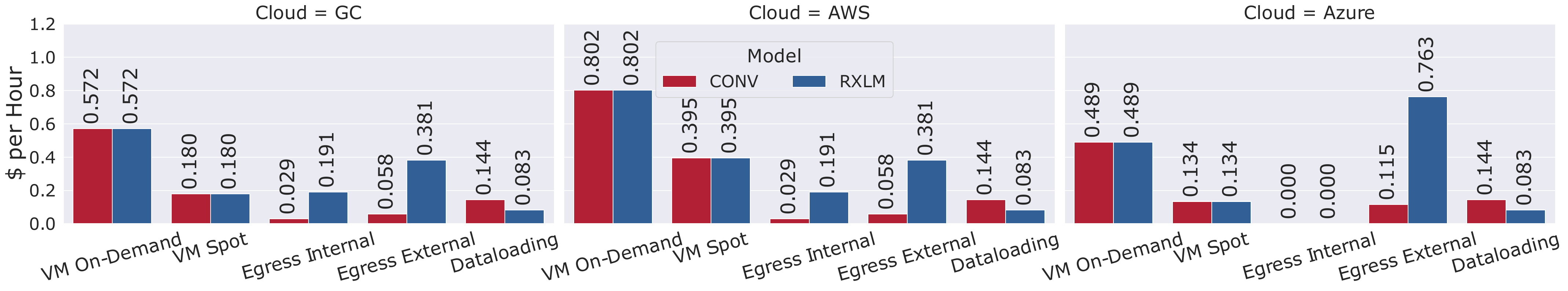}
        \vspace{-6mm}
        \caption{Intra- and inter-zone in the US region (D-2/3).}
        \label{fig:multi-cloud-costs-d3}
    \end{subfigure}
    \begin{subfigure}[c]{0.47\textwidth}
        \includegraphics[width=\textwidth]{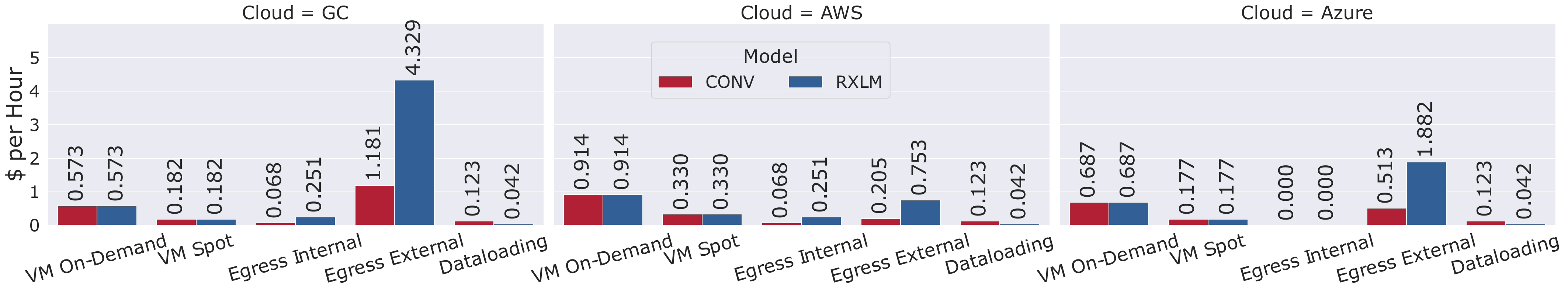}  
        \vspace{-6mm}
        \caption{Intercontinental in the US, EU, ASIA and AUS (C-8).}
        \label{fig:multi-cloud-costs-c8}
    \end{subfigure} 
    \vspace{-3mm}
    \caption{Costs breakdown for D-2/3 and C-8 experiments.}  
    \label{fig:normalized-multi-cloud-costs}
    \vspace*{-6mm}
\end{figure}  

\textbf{(2) External egress costs can overshadow VM costs.} One drawback to training in multiple regions or zones is that egress traffic can incur additional costs depending on the cloud provider.
We have summarized the cost of egress traffic within a zone (intra-zone), between zones in each region (inter-zone), and between continents in ~\Cref{tab:cloud-pricing}. 
Notably, any traffic to Oceania (Australia, New Zealand, and others, abbreviated as OCE) generates the highest cost of \$0.15/GB for GC.
We have broken down the costs for the multi-cloud experiment in ~\Cref{fig:multi-cloud-costs-d3} on an hourly per-VM basis. 
With only four peers in the D-1/2/3 experiments, we have an N-to-N communication, i.e., each peer sends its gradients to every other peer.
This means that $\frac{1}{3}$ of the egress was internal to the partner VM in the same cloud, and the remaining $\frac{2}{3}$ went to the remaining two peers in the other cloud.

First, loading data from Backblaze costs \$0.01/GB from anywhere in the world, which gives us a rate of \$0.144/h for the CV and \$0.083/h for the NLP experiments.
Even when CV throughput is less than half of the NLP model (\Cref{fig:multi-cloud-throughput}), images are much larger than text, resulting in a higher data rate.
While this is close to the spot instance costs of GC (\$0.18/h) and Azure (\$0.134/h), these are one-time costs until the entire dataset is downloaded and retrieved from the disk cache, assuming large enough local storage.
A more detailed comparison of cloud provider storage offerings is beyond our scope, but current prices range from \$0.02/GB to \$0.14/GB in various GC regions, making our setting (B2) competitive.

Second, the external egress costs for the NLP experiments are very high compared to the other costs.
They are 2.2x higher than the spot instance for GC and 5.7x higher for Azure, as the traffic costs in the US zone are \$0.01/GB and \$0.02/GB, respectively.
The Azure cost is even higher (\$0.763/h) than the on-demand instance price of \$0.489/h.
The CV experiments are much less affected due to the smaller model size, but Azure still manages to almost match its spot instance price of \$0.134/h with the external egress cost of \$0.115/h.

Finally, the total compute cost, including egress and data loading in this multi-cloud constellation, is the sum of all the cloud providers' prices times the number of VMs used.
For the CV experiments, GC, AWS, and Azure cost \$0.762/h, \$1.192/h, and \$0.363/h, respectively, making the combination of GC with Azure 42\% cheaper than GC with AWS.
For the NLP experiments, GC, AWS, and Azure cost \$0.835/h, \$1.05/h, and \$0.973/h, respectively, and GC combined with Azure is better than GC with AWS by a smaller margin of 3.9\%.
However, the intercontinental network egress prices for both GC and Azure are up to 15 times higher than the inter-zone prices, so what about the cost-effectiveness compared to geo-distributed experiments?

\textbf{(3) Geo-distributed egress can incur most of the cost.} To illustrate the cost of intercontinental training, we use our C-8 experiment with two VMs in four continents from ~\Cref{sec:geodistributed-performance} to plug the cost for each cloud provider.
The egress costs are calculated slightly differently than in the D-2 and D-3 experiments because four groups of two VMs average locally and then distribute the gradients across the other groups.
This results in $\frac{8}{20}$ internal egress calls (two calls between each group), $\frac{6}{20}$ intercontinental egress calls (two calls between three regions), and $\frac{6}{20}$ AUS egress calls (three regions share their gradients with AUS and vice versa).

~\Cref{fig:multi-cloud-costs-c8} shows the resulting egress traffic cost per VM.
The high cost between continents scales to a multiple of the remaining cost for CV and NLP with GC and Azure.
For NLP, the external egress cost for GC is \$4.329/h, more than 90\% of the total cost per VM (\$4.804/h).
Even with Azure having a more moderate rate of \$0.02/GB for intercontinental communication and only \$0.08/GB for OCE traffic, it still results in \$1.882/h external egress cost (\$2.101/h total).
This is in contrast to AWS, which has a cap of \$0.02/GB to any location, resulting in the best total cost of \$1.376/h per VM.
The relatively high AWS instance cost compares favorably to the other cloud providers regarding geo-distributed training.
Keeping egress traffic in mind when deciding to scale to other continents is essential, as it can be the most significant part of the total cost.
This raises another question: If egress traffic matters so much, how does model size affect it?
\begin{figure}
    \begin{subfigure}[c]{0.225\textwidth}
        \includegraphics[width=\textwidth]{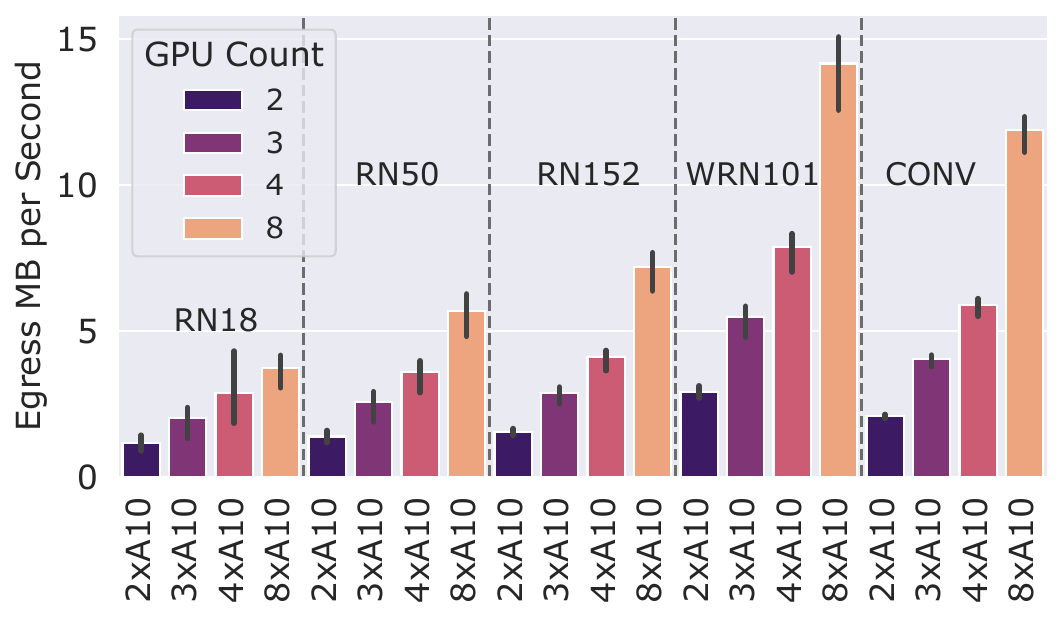}
        \vspace{-4mm}
        \caption{CV}
        \label{fig:cv-baseline-egress}
    \end{subfigure}
    \begin{subfigure}[c]{0.23\textwidth}
        \includegraphics[width=\textwidth]{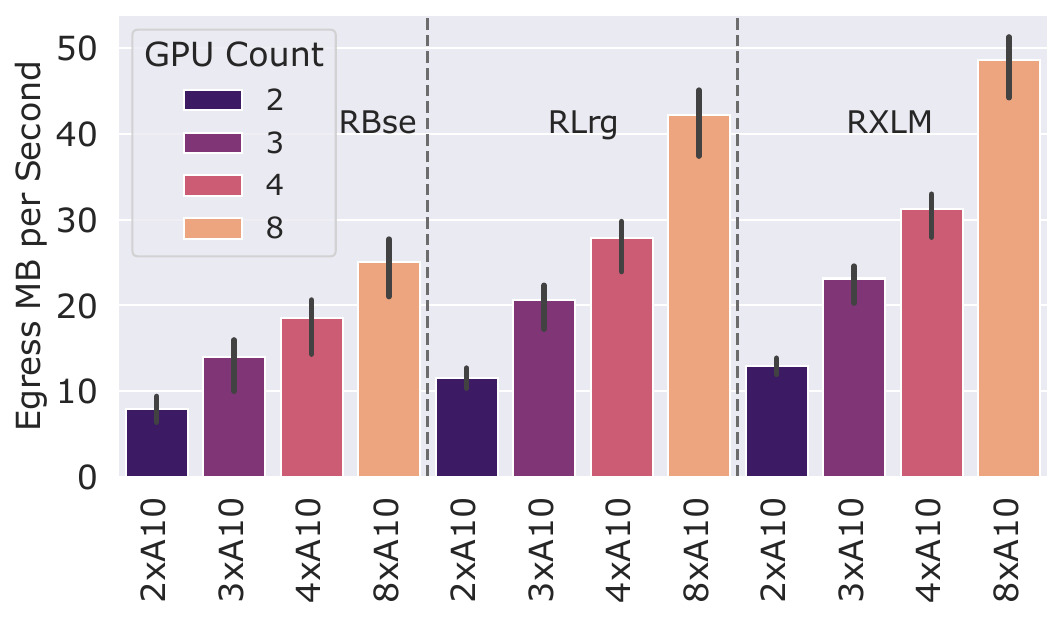} 
        \vspace{-4mm}
        \caption{NLP}
        \label{fig:nlp-baseline-egress} 
    \end{subfigure}
    \vspace{-3mm}
    \caption{Baseline egress rate on 2-8 A10 GPUs.}
    \label{fig:baseline-egress}
    \vspace*{-4mm} 
\end{figure} 
 
\textbf{(4) Small models have lower egress rates than larger models.} Model size affects two parts of the distributed training time.
First, larger models tend to have slower averaging rates, but more data movement costs due to their size. However, larger models are also averaged less frequently because they take longer to perform a step.
To analyze this, we review the experiments in ~\Cref{sec:model-suitability}, where we evaluate different model sizes and GPUs counts.
~\Cref{fig:baseline-egress} shows the average egress rate over each experiment's runtime for both CV and NLP from two to eight A10 GPUs.
The trend is clear: the smaller the model, the lower the egress rate for all GPUs (e.g., RN18 vs. RN50).
This is surprising, as the "square-cube" law~\cite{ryabinin2023swarm} states that with a decrease in parameters, the calculation time will decrease quadratically while the communication time decreases linearly.
This means that with a sufficiently small model, most of the training will consist of communication time, and the egress rate would increase, as it is defined through $\frac{\text{parameter count}}{\text{calculation time}}$.
However, we find that even with our smallest model, RN18, with 11.7M parameters and eight A10 GPUs, we are still not at the point where the communication time takes up most of the time.

In summary, multi-cloud training is generally possible and can be cost-effective when keeping the egress costs and granularity in mind.
Regardless of the cloud provider, staying in the same region is preferred, with the US having the most favorable egress price offers.
A significant portion of the cost may be hidden in egress costs, accounting for more than 90\% of the total cost in our NLP experiments in GC and Azure.
Based on the additional egress costs alone, renting on-demand hardware may be more advantageous than using spot instances between different regions.
CV training is generally more calculation- than communication-heavy, resulting in slightly higher data-loading but fewer egress costs.
However, from our experiments, this is a favorable trade-off because data-loading is much cheaper than egress costs.

\section{Hybrid-cloud Performance}
\label{sec:hybrid-cloud-performance}
\begin{table}[h]
    \caption{Average hybrid-cloud throughput and latency.}
    \vspace*{-3mm}
    \scalebox{0.60}{
    \begin{subtable}[h]{0.35\textwidth}
        \centering
        \caption{Single stream TCP throughput in Gb/s.}
        \begin{tabular}{l|r|r|r}
        \backslashbox{\textbf{From}}{\textbf{To}} & \textbf{EU T4} & \textbf{US T4} & \textbf{US A10} \\ \hline
        \textbf{RTX8000}  & \cellcolor[HTML]{fadc24}0.45 & \cellcolor[HTML]{fa7e24}0.06 & \cellcolor[HTML]{fa7e24}0.05 \\ \hline
        \textbf{DGX-2 (8xV100)} & \cellcolor[HTML]{fadc24}0.55 & \cellcolor[HTML]{fa7e24}0.08 & \cellcolor[HTML]{fa7e24}0.07 
        \end{tabular}
        \vspace{1mm}
        \label{tab:local-experiments-bandwidth}
    \end{subtable}
    } 
    \scalebox{0.65}{ 
    \begin{subtable}[h]{0.35\textwidth}
        \centering   
        \caption{ICMP Latency in ms.}
        \begin{tabular}{l|r|r|r}
        \backslashbox{\textbf{From}}{\textbf{To}} & \textbf{EU T4} & \textbf{US T4} & \textbf{US A10} \\ \hline
        \textbf{RTX8000}  & \cellcolor[HTML]{fadc24}16.73 & \cellcolor[HTML]{fa7e24}150.80 & \cellcolor[HTML]{fa7e24}159.05 \\ \hline
        \textbf{DGX-2 (8xV100)} & \cellcolor[HTML]{fadc24}16.19 & \cellcolor[HTML]{fa7e24}150.27 & \cellcolor[HTML]{fa7e24}158.54
        \end{tabular}
        \vspace{1mm}
        \label{tab:local-experiments-ping}
    \end{subtable}
    }
    \label{tab:local-experiments-network}
    \vspace*{-5mm}
\end{table} 
Can augmenting on-premise hardware with cloud resources be worthwhile to speed up DL training?
In this section, we examine two settings: \textbf{(E)}, where a consumer-grade GPU, the RTX8000, is deployed on-site, and \textbf{(F)}, where a server-grade node, the DGX-2 (8xV100), is deployed on-site. We vary the extra resources, between one to eight T4 EU (\{E,F\}-A), T4 US (\{E,F\}-B) and A10 US (\{E,F\}-C) GPUs.

\textbf{Experimental design.} In both settings, we want to investigate how to extend local hardware with cloud resources and when this leads to better throughput.
The cloud resources, in this case, are the same US/EU GC T4 instances as in ~\Cref{sec:geodistributed-performance} and the US LambdaLabs A10 GPUs from ~\Cref{sec:model-suitability}.
We double the number of cloud VMs with each increment, starting with one additional GPU (i.e., E-A-1) until we have eight additional cloud VMs (i.e., E-A-8). 
This allows us to compare the same hardware in the EU and the US, and slightly weaker, local hardware (EU T4) and better, but more distant hardware (US A10).

Both the \textbf{(E)} and \textbf{(F)} setups share the network uplink between 450 and 550~Mb/s to the EU datacenter in Belgium, as they are located in the same building in Europe (\Cref{tab:local-experiments-network}). 
However, as this is not a Google-owned datacenter, the traffic is partly going over the public internet, which results in a lower bandwidth of 50 and 80~Mb/s to the US-based VMs compared to 210~Mb/s between the US and EU GC datacenters (\Cref{tab:geodistributed-throughput}).

\begin{table}[]
\begin{center}
\caption{Hybrid- vs. cloud-only throughput for the \textbf{(E)} setting.}
\vspace*{-2mm}
\scalebox{0.8}{
\begin{tabular}{r|r|r|r|r|r|r}
 \backslashbox{\textbf{Model}}{\textbf{Setup}} & \textbf{RTX8000} & \textbf{E-A-8} & \textbf{E-B-8} & \textbf{E-C-8} & \textbf{8xT4} & \textbf{8xA10} \\ \hline
CONV &   \cellcolor[HTML]{f17d3a}194.8 & \cellcolor[HTML]{f1be3a}316.8 & \cellcolor[HTML]{f1be3a}283.5 & \cellcolor[HTML]{c3f13a}429.3 & \cellcolor[HTML]{f1be3a}261.9 & \cellcolor[HTML]{67F13A}620.6 \\ \hline
RXLM    &   \cellcolor[HTML]{f1be3a}431.8 & \cellcolor[HTML]{c3f13a}556.7 & \cellcolor[HTML]{f1be3a} 330.6 & \cellcolor[HTML]{f17d3a}223.7 & \cellcolor[HTML]{c3f13a}575.1 & \cellcolor[HTML]{67F13A}1059.9     
\end{tabular}
}
\end{center}
\label{tab:hybrid-vs-cloud-only-startup}
\vspace*{-3mm}
\end{table}

\begin{figure}
    \begin{subfigure}[c]{0.24\textwidth}
        \includegraphics[width=\textwidth]{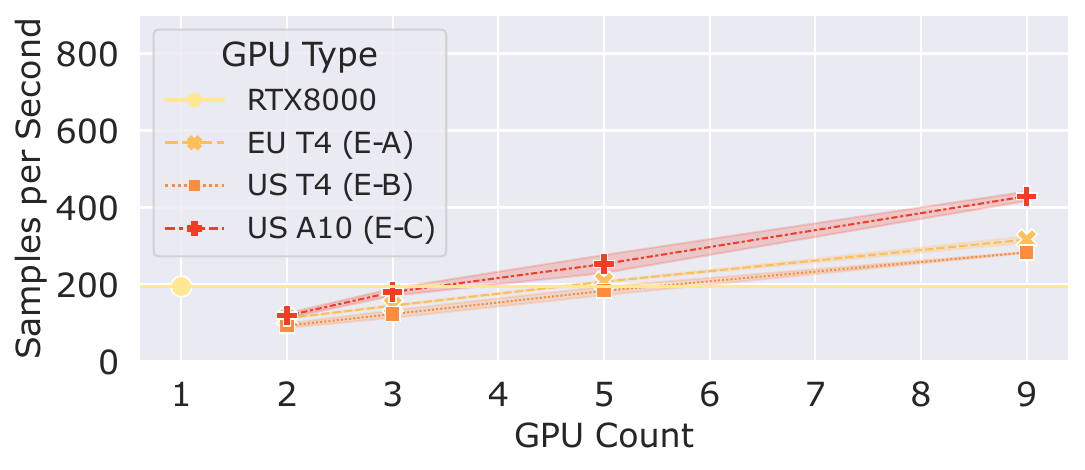}
        \vspace{-6mm}
        \caption{CV Throughput}
        \label{fig:cv-private-hybrid-cloud-throughput}
    \end{subfigure}
    \begin{subfigure}[c]{0.21\textwidth}
        \includegraphics[width=\textwidth]{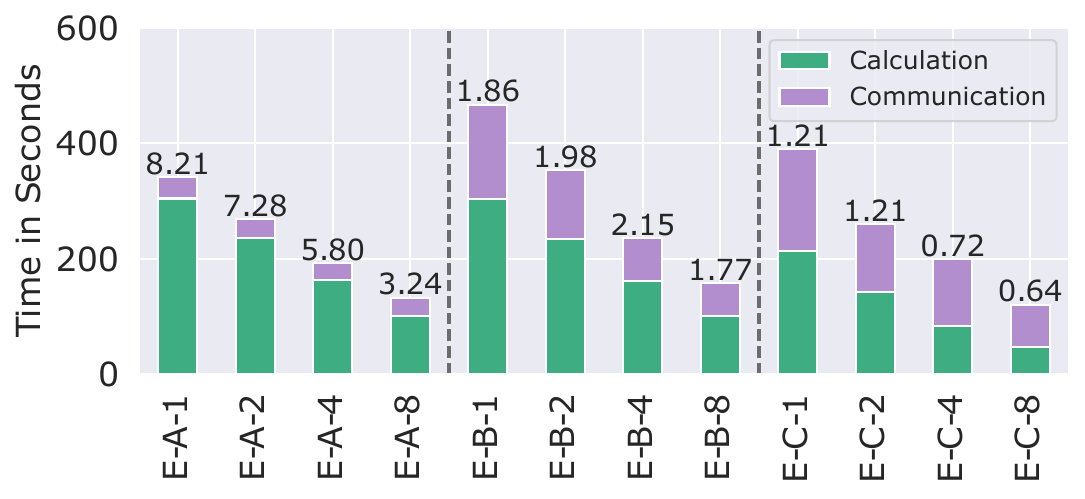} 
        \vspace{-4.7mm}
        \caption{CV Granularity}
        \label{fig:cv-private-hybrid-cloud-granularity}
    \end{subfigure}
        \begin{subfigure}[c]{0.24\textwidth}
        \includegraphics[width=\textwidth]{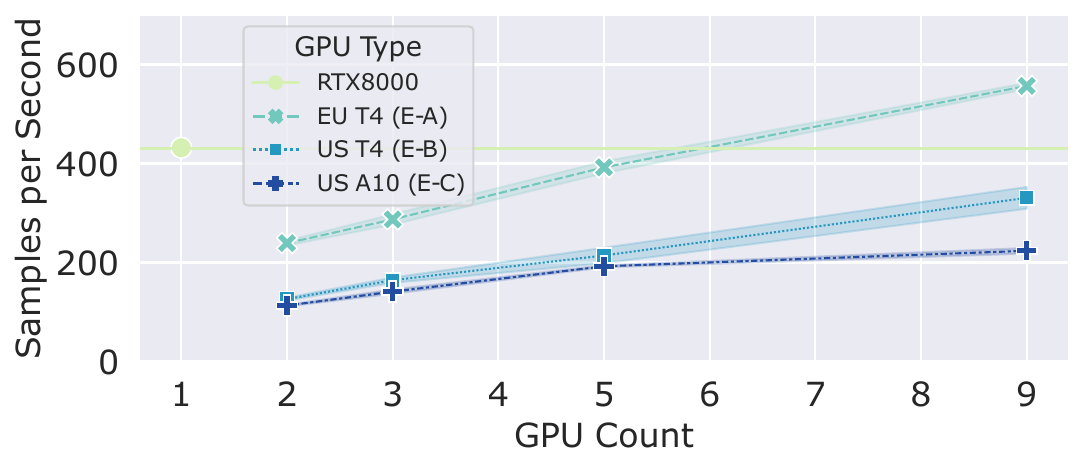}
        \vspace{-6mm}
        \caption{NLP Throughput}
        \label{fig:nlp-private-hybrid-cloud-throughput}
    \end{subfigure}
    \begin{subfigure}[c]{0.21\textwidth}
        \includegraphics[width=\textwidth]{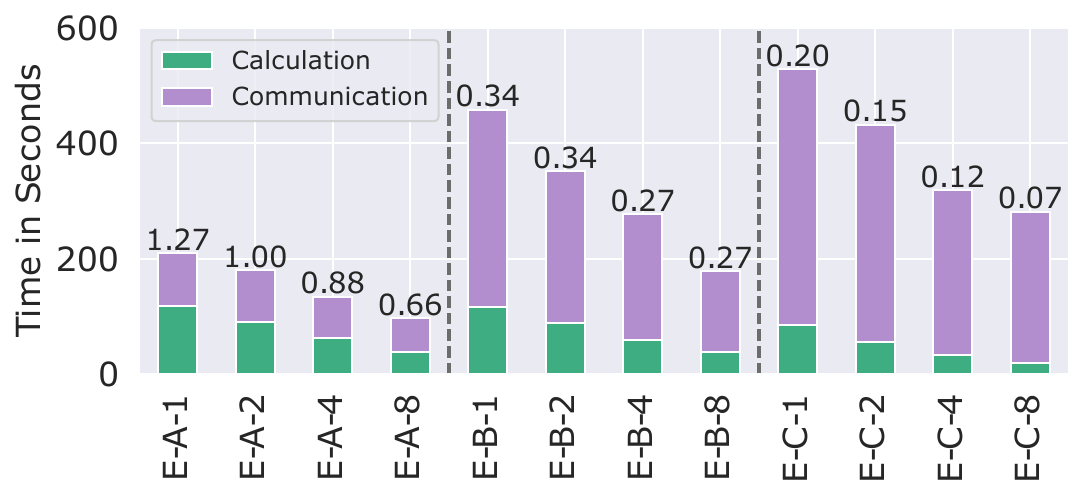}  
        \vspace{-4.7mm}
        \caption{NLP Granularity} 
        \label{fig:nlp-private-hybrid-cloud-granularity}
    \end{subfigure}
    \vspace{-3mm}
    \caption{Hybrid-cloud experiments for the \textbf{(E)} setting.}
    \label{fig:private-hybrid-cloud-performance}
    \vspace*{-2mm}
\end{figure} 

\textbf{(E) Consumer-grade setting.} The results follow the same trend as in ~\Cref{sec:geodistributed-performance}.
The CV task has a higher granularity of 8.21 with 2 GPUs at E-A-1 than NLP (1.27) (\Cref{fig:cv-private-hybrid-cloud-granularity,fig:nlp-private-hybrid-cloud-granularity}), and scales regardless of the location of the cloud resources (\Cref{fig:cv-private-hybrid-cloud-throughput}).
We almost match the baseline throughput of 195 SPS at 5 GPUs in all settings for CV (E-A-4, E-B-4, E-C-4).
The best throughput was reached at E-C-8 with the US A10 GPUs with 429 SPS.
For NLP, only the E-A-8 experiment beats the baseline with a speedup of 1.29x and 556 SPS due to the low granularity and the intercontinental base penalty for the US experiments.

However, is combining on-premise and remote cloud resources better than using the cloud without paying the intercontinental bandwidth tax?
To analyze this, we compare the \textbf{(E)} experiments with the 8xA10 experiment from ~\Cref{sec:model-suitability} and 8xT4 experiment from ~\Cref{sec:geodistributed-performance} in ~\Cref{tab:hybrid-vs-cloud-only-startup}.
First, the 8xA10 experiments are the fastest for both CV and NLP, which removes the respective hybrid-cloud combination from contention (E-C-8).
Second, the 8xT4 experiments for NLP are faster than any other hybrid-cloud setup, making the cloud-only solution favorable.
Finally, while we always beat the baseline 8xT4 CV throughput (261.9 SPS), but in the case of E-B-8 (283.5 SPS), just barely.
The throughput of E-A-8 (316.8 SPS) makes the hybrid-cloud setup the most favorable in terms of relative GPU scaling (32.5 SPS per GPU), but it does not come close to the best cloud-only throughput of 8xA10 with 620.6 SPS.

Summarizing, the cloud-only experiments are the fastest overall due to their single-GPU throughput and locality.
Adding cloud resources to on-premise hardware leads to a high communication time, which is not compensated by the additional processing speed of the GPUs.
Proximity to the on-premise hardware is essential, as the more local cloud resources (E-A-8) consistently resulted in a better throughput than the same remote cloud resources (E-B-8).

\begin{figure}
    \begin{subfigure}[c]{0.24\textwidth}
        \includegraphics[width=\textwidth]{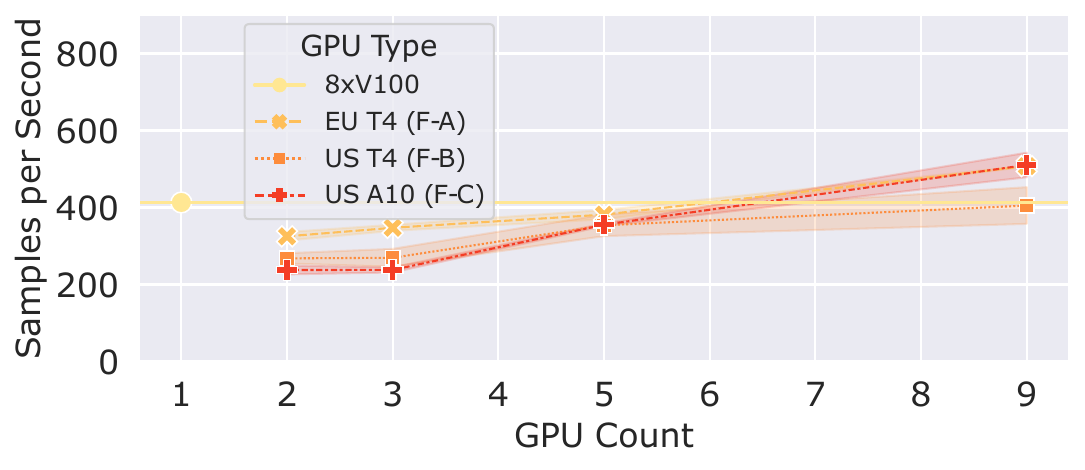}
        \vspace{-6mm}
        \caption{CV Throughput} 
        \label{fig:cv-research-hybrid-cloud-throughput}
    \end{subfigure}
    \begin{subfigure}[c]{0.21\textwidth}
        \includegraphics[width=\textwidth]{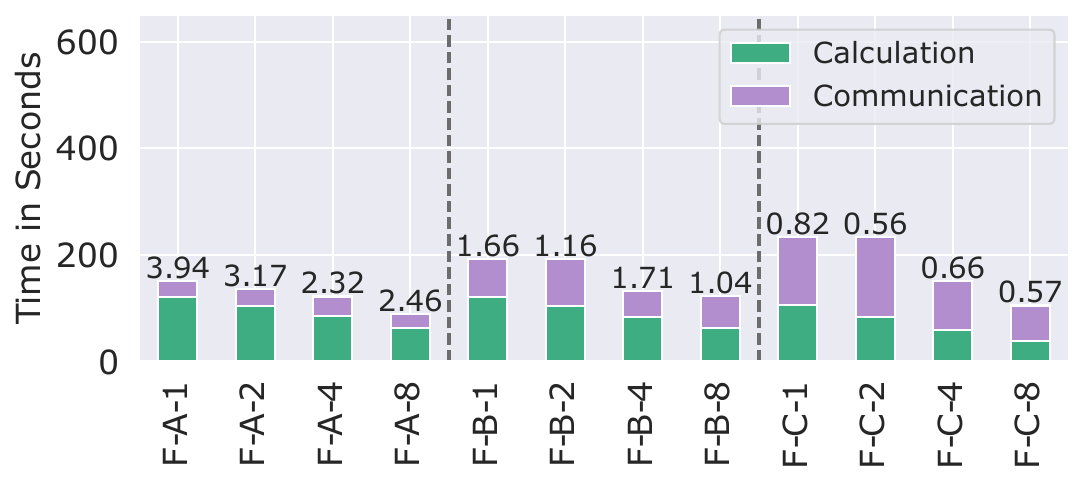}  
        \vspace{-4.7mm}
        \caption{CV Granularity}
        \label{fig:cv-research-hybrid-cloud-granularity}
    \end{subfigure}
        \begin{subfigure}[c]{0.24\textwidth}
        \includegraphics[width=\textwidth]{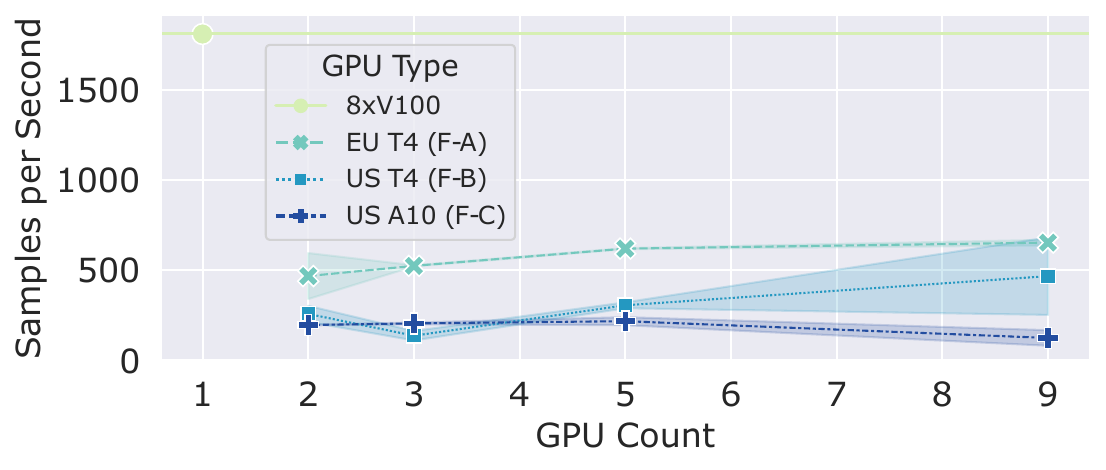}
        \vspace{-6mm} 
        \caption{NLP Throughput}
        \label{fig:nlp-research-hybrid-cloud-throughput}
    \end{subfigure}
    \begin{subfigure}[c]{0.21\textwidth}
        \includegraphics[width=\textwidth]{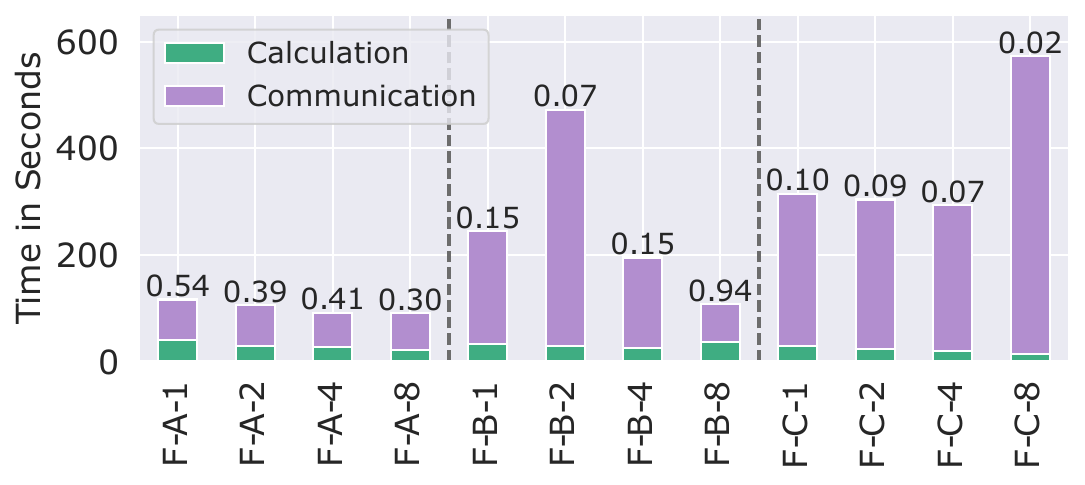}  
        \vspace{-4.7mm} 
        \caption{NLP Granularity} 
        \label{fig:nlp-research-hybrid-cloud-granularity}
    \end{subfigure}
    \vspace{-3mm}
    \caption{Hybrid-cloud experiments for the \textbf{(F)} setting.}
    \label{fig:research-hybrid-cloud-performance}
    \vspace*{-2mm}
\end{figure} 

\textbf{(F) Server-grade setting.} The baseline throughput is significantly higher compared to the RTX8000, with a much more powerful 8xV100 DGX node to 413~SPS for CV and 1811~SPS for NLP (\Cref{fig:cv-research-hybrid-cloud-throughput,fig:nlp-research-hybrid-cloud-throughput}) via {PyTorch \color{diff1}data parallelism~\cite{li2020pytorch}}.
This increases the penalties from ~\Cref{sec:model-suitability}, leading to the only speedup from baseline for CV in experiments F-A-8 (507~SPS) and F-C-8 (510~SPS).
This is surprising, as the older T4 GPUs in the EU perform similarly to the much newer A10 GPUs in the US, showcasing the trade-off between slower, local compute and faster, remote compute.
The granularity of 2.46 for F-A-8 shows that there is enough calculation time to distribute, while the F-C-8 experiments spend $\approx 62\%$ of the total training time on communication with a granularity of 0.57 (\Cref{fig:cv-research-hybrid-cloud-granularity}).
The NLP experiments never reach the baseline throughput of the 8xV100 due to using most of the time for communication. 
The NLP F-B and F-C experiments mainly consist of communication (\Cref{fig:nlp-research-hybrid-cloud-granularity}) with a granularity of up to 0.02, which results in a nonlinear, unstable training time due to the minimum matchmaking time issue \textbf{(2)} from ~\Cref{sec:model-suitability}.

In summary, the hybrid-cloud experiments conclude that while on-premise hardware can be augmented with cloud resources, it will likely be cost-efficient if all resources are on the same continent.
Using only cloud resources is more advantageous if the on-premises hardware is not co-located.

\section{Further Insights}
\label{sec:further-insights}

\textbf{Communication time can decrease with more peers.}
Let us compare the granularity of the experiments for E-B (\Cref{fig:cv-private-hybrid-cloud-granularity}), which uses T4 GPUs in the US as an additional cloud resource.
\textit{Both} the computation and communication time decrease with the number of GPUs, even increasing the granularity from 1.98 at E-B-2 to 2.15 at E-B-4.
This is surprising since, usually, with more peers, the communication time should increase, and the US-EU communication bottleneck should slow us down to the same extent as the E-B-1 experiment.
This reduction is a Hivemind-specific anomaly, as it uses a single TCP stream per peer.
With TCP, there needs to be an acknowledgment (ACK) of each packet by the receiving peer, which is impacted by the connection's latency.
In our high latency network between continents, the round trip time (RTT) of 300-318ms limits the maximum bandwidth a single TCP stream to 50-80~Mb/s.
However, a way to improve link utilization is to use multiple streams, one for each peer, which we encounter in experiments E-(B|C)-2,4,8.
To verify the potential gains, we perform a microbenchmark of the multi-stream bandwidth from the RTX8000 to the EU and US data centers.
Although there is wide variation, likely due to network utilization, with 80 clients, we achieve a maximum bandwidth of 6~Gb/s within the EU and up to 4~Gb/s to the US. 
While larger peer groups and, consequently, larger models benefit from multi-peer communication by default and do not see significant changes in communication time, small models in unevenly distributed VMs setups can be disproportionately affected.
The same trend can be observed in all high latency experiments (i.e., between the EU and the US), e.g., E-B, E-C for CV and NLP (\Cref{fig:cv-private-hybrid-cloud-granularity,fig:nlp-private-hybrid-cloud-granularity}, and F-B and F-C for CV (\Cref{fig:cv-research-hybrid-cloud-granularity}).
In summary, uneven distribution of computational resources in high-latency networks (e.g., intercontinental) can reduce communication time with Hivemind due to more parallelism, lessening the impact of low bandwidth for a single data stream.

\textbf{Cost analysis.} The DGX-2 (8xV100) node from ~\Cref{sec:hybrid-cloud-performance} represents server-grade hardware that could be used to train models.
However, how does it compare in throughput per \$ to all of our distributed cloud experiments?
\begin{figure}
    \includegraphics[width=0.45\textwidth]{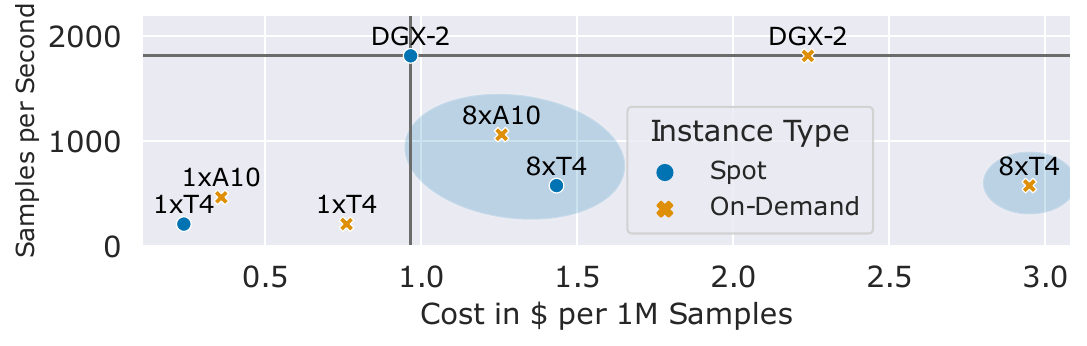}
    \vspace{-3mm}
    \caption{Cost to throughput tradeoff for RoBERTaXLM at different instance types. Our training setups (circled), that are due the low granularity of the NLP model, neither cheaper, nor faster than the centralized offering (DGX-2).} 
    \label{fig:nlp-sps-trade-off}
    \vspace*{-2mm}
\end{figure}
The ~\Cref{fig:cv-sps-trade-off} (CV) and ~\Cref{fig:nlp-sps-trade-off} (NLP) show the complete cost analysis of the DGX-2, the 8xT4 experiments, and the 8xA10 experiments for spot and on-demand pricing. We use the internal egress costs from ~\Cref{fig:multi-cloud-costs-d3} as a reference for the 8xT4 setup.
For simplicity, we compare the spot pricing without interruptions, as we assume that a new VM can be spun up fast enough not to affect the training throughput in the long run.
We mark the centralized baseline (DGX-2) cost per 1M samples and the throughput in samples per second with a horizontal and vertical line.
This means that we are cheaper to the left to the vertical line, and above the horizontal line, we are faster (and vice versa).
We circle the new value propositions that we enable in both figures.
{\color{diff1} Our hardware setups have additional key characteristics: They are resilient by default to interruptions due to running in a decentralized fashion and they enable the combination of more GPUs than cloud providers offer in a single node.
Currently, common hardware configurations (DGX) allow up to eight GPUs connected via NVLink, and with older hardware, only up to 4xT4s connected via PCIe at 10\;GB/s between GPUs (with GC).
We were able to combine eight single GPU nodes from GC and LambdaLabs to create competing performance and price setups without dedicated GPU interconnects.
}

A spot DGX-2 costs at the time of writing \$6.30/h (\$14.60/h on-demand) in GC US, which makes it the best value proposition for the low granularity NLP task.
It is followed by the 8xA10, which are 41\% slower and 30\% more expensive than the DGX-2 (\Cref{fig:nlp-sps-trade-off}). 
The 8xT4 experiments are even more expensive, as the internal egress costs take up more than half of the costs, making them the worst value proposition.
However, for CV, we manage to provide two new offerings: First, the 8xA10, which is both 50\% faster and 49\% cheaper than the DGX-2, and 8xT4, which is 58\% cheaper than DGX-2, while being 37\% slower (\Cref{fig:cv-sps-trade-off}).
The CV model can be scaled more easily due to its initially high granularity, which makes the very competitive offering of \$0.6/h per A10 from LambdaLabs an excellent value proposition.
However, while we only evaluated eight T4 GPUs for our GC-based experiments, with a granularity of 5.19 (CV A-8 in ~\Cref{fig:geo-dist-us-only-granularity}), there is ample space to scale even further.
It is important to note that LambdaLabs does not charge for any data egress, but GC does with \$0.01/GB, and the 8xT4 experiment is still cheaper.  
While LambdaLabs is often at capacity, Google Cloud positions itself as a hyperscaler with the advantage of rarely being at max occupancy.

{\color{diff1}
We also evaluated the performance of the 4xT4 PyTorch DDP~\cite{li2020pytorch} for CV with the best available multi-T4 node on GC (4xT4).
The NLP experiments ran OOM.
Since the DDP 4xT4 runs on a single node, it causes no interconnect costs and is priced at \$0.96 per 1M samples at spot pricing, while our 8xT4 setup costs \$1.77 per 1M samples (84\% more expensive).
However, the 8xT4 setup has a higher throughput of 262 SPS (26\% faster) compared to the 4xT4 node (207 SPS).
This higher speed is not available at the price point of the 4xT4 node.
Moreover, the 8xT4 setup has the potential for further scaling, which we discussed in detail in \Cref{sec:geodistributed-performance}.
}

In summary, the lower spot prices for older GPUs allow us to train models more cost-efficiently when task granularity allows it and get more value per \$ when training on the 8xT4 or 8xA10 compared to an DGX-2 node.
{\color{diff1}
Combining multiple nodes with single GPUs with lower bandwidths enables scaling that was previously impossible to achieve without resorting to much more powerful GPUs.
}
Distributed spot instance pricing opens up a new value proposition compared to on-demand offerings that can even compete with the competitive pricing of smaller cloud providers.

{\color{diff}
\textbf{Spot VM Interruption Frequency.}
While we used low spot prices as a cost-saving argument in our experiments, we did not elaborate on the most significant drawback - the possibility of being terminated by the cloud provider at any time. 
There is already some research on how different cloud providers track the interruption frequency and can be used for varying workloads to achieve a positive \$-per-throughput effect~\cite{9975369,yang2022schedulingml,yang2023skypilot}.

Interruption affects three aspects: First, the interruption frequency is defined by AWS as the number of VMs terminated in the last 30 days, which is between 5 and 20\%~\cite{awsspotprices}.
This value was not representative during our experiments with any cloud provider, as we noticed that it is highly dependent on the time of day of the zone. 

Second, the time needed to setup a VM until training starts.
The startup time of a VM depends on the cloud provider (e.g., a preconfigured image) and the one's technology stack (e.g., Docker, Kubernetes, Ansible).
In our experience, VM startup time ranges between seconds to minutes with manual deployment taking up to 10 minutes.
Although startup time can be improved, model training typically takes multiple hours or days, making it a less impactful optimization.

Third, the time required for the new peer to synchronize the training state with other peers.
In our experience, this took at worst two hivemind epochs due to the averaging starting before synchronization is finished.
While it is possible to create a hivemind epoch that is short enough to prevent new peers from joining, this only happens with a low enough granularity where scaling is not beneficial anymore as we are mostly communication bound.

Finally, while the VM setup and synchronization of the training state take time, the interruption frequency significantly affects the final throughput. We faced difficulties acquiring even a single spot VM during our GC experiments during daylight hours.
This highlights the need for systems like SkyPilot~\cite{yang2023skypilot}, which utilizes automation to deploy spot instances across various clouds and zones.
In our case, the interruption frequency can be used as a penalty on the training throughput, i.e., a 5\% interruption frequency over the entire training time means roughly a 5\% slower training.
}
\vspace*{-8mm}
\section{Lessons Learned}
\label{sec:lessons-learned}
We find it important to summarize our findings more generically to provide guidance for DL practitioners that want to perform distributed spot training. These lessons are based on the ~\Cref{sec:model-suitability,sec:geodistributed-performance,sec:multicloud-performance,sec:hybrid-cloud-performance}.

\textbf{Small model training still scales.}
We have shown that models between 12M-560M parameters can be trained in a decentralized, distributed fashion achieving a speedup of up to 4.37x on eight Ampere-GPUs.
The limiting factor as to when a model is suitable for (geo-)distributed training is the target batch size which all peers need to accumulate until synchronization happens.
We found a TBS of 32K suitable to not only train in a single zone, but even see a speedup when using VMs in four different continents.
As long as the optimizer can handle big-batch training and the dataset is big enough to accommodate large batches, the remaining issue to find the base granularity of the model to decide how to scale it cost-effectively.
Finally, we found that small models induce less traffic over larger models over time, even at a much higher averaging rate, making them better suited for cost-efficient training than large models.

\textbf{Egress costs can take up most of the total cost.}
Egress pricing for the NLP experiments overtook the spot and the on-demand costs of T4 GPUs when training on four continents or even in two zones.
For example, RoBERTaXLM's high throughput and parameter count require more data to be sent between peers during averaging due to smaller granularity.
Under the current pricing models, AWS has the best value for geo-distributed training, while GC and Azure are best at training in a single zone.
The biggest cost-saving potential lies in cloud providers that do not charge for egress at all, like LambdaLabs.



\textbf{Granularity is important to evaluate scalability.}
We found that the ratio between calculation and communication time, granularity, is the most important metric to track when deciding on distributed training suitability.
It enables us to compare the scalability potential between different models on the same hardware due to summarizing their model size and throughput ratio.
Additionally, it gives a value to the cost-efficiency: With a granularity of exactly 1, the potential speedup when doubling the number of VMs is, at best, 1.33x due to halving the calculation time.
However, with a granularity of 10, the speedup with double the VMs is, at best, 1.83x due to the communication time playing a less significant role.
With this, we can estimate training performance with additional resources.

\textbf{Geo-distributed multi-cloud training is possible and is cost-efficient.}
Even with the current teething pains of Hivemind, we got a speedup in all of our experimental setups of intra-zone, transatlantic, and intercontinental settings as long as the granularity of the task permitted it.
Using older and cheaper Tesla GPUs at spot pricing is not only more cost-efficient than the DGX-2 offering, but even trumps the competitive pricing model of LambdaLabs, all while including egress costs.
Our network profiling showed that the current training limitations are not primarily the bandwidth but rather the intercontinental latency and the task's granularity.
If the granularity is already low at high bandwidth, it can only worsen when used in a high latency, low bandwidth network.
When considering both, estimating the potential cost-savings of investing in a multi-/hybrid-cloud scenario is possible.

\vspace*{-2mm}
\section{Related Work}
\label{sec:related-work}
\textbf{Decentralized deep learning.}
Training with unreliable peers has been studied in a collaborative setting, resulting in the Distributed Deep Learning in Open Collaborations (DeDLOC)~\cite{diskin2021distributed} algorithm, on which the Hivemind framework~\cite{hivemind} is based.
It can interpolate between traditional distributed DL algorithms like parameter servers~\cite{li2014scaling}, decentralized SGD~\cite{lian2017can}, or All-Reduce SGD~\cite{sergeev2018horovod}.
We used the Hivemind framework for all of our experiments, as it provided the base for training on spot instances in high latency, low bandwidth networks.

SWARM~\cite{ryabinin2023swarm} applies both previous techniques and adds model parallelism to the mix by creating pipelines between nodes and rebalancing them in case of failures.
The authors find a crucial insight in the "square-cube" law, which argues for better training scalability with larger model sizes; as the size increases linearly, so does the communication time, while the calculation time increases quadratically.
We add to that by analyzing distributed training for smaller model sizes that pose different trade-offs. 
We show that while the square-cube law still holds for increasing model sizes, under consideration of granularity, we can still train small models.

{\color{diff}
Decentralized deep learning on heterogeneous hardware with slow interconnects can benefit the training of foundation models. To achieve this, model and pipeline parallelism can be used in addition to data-parallel training~\cite{yuan2022decentralized}.
This is a complementary work to ours, since we target smaller models and weaker hardware.
}

\textbf{Deep learning on spot instances.}
DeepSpotCloud~\cite{lee2017deepspotcloud} is a system that uses the AWS API to automatically migrate a DL task with checkpointing whenever the spot instance is terminated.
The authors note that the volatility of GPU instance pricing and interruptions have a unique pattern compared to non-accelerated VMs, and solve this by using intercontinental provisioning.
We noticed the same trends of high interruption ratios in our experiments.
However, we have shown that geo-distributed training is possible until granularity permits it, which poses a possibility for ever-migrating training between continents without checkpointing. 

Amazon Sagemaker~\cite{das2020sagemaker} is an AWS service that allows to perform ML under budget constraints.
For training, it supports spot VM migration until a cost threshold is reached by checkpointing the progress.
{\color{diff}
However, it lacks the option of training on multiple spot VMs.
It can do either spot instance training on DGX-like nodes or combine multiple on-demand nodes with PyTorch DDP (or similar), but not both.
This eliminates the potential of accelerating the training process with more GPUs that do not fit a single spot-provisioned hypervisor.
}

The analysis by Yang et al.~\cite{yang2022schedulingml} investigates maximizing a target accuracy from a spot pricing versus time perspective.
Linear programming was used to decide how to provision the VMs with different cost-utility trade-offs.
While this shows the potential of utilizing multiple clouds and continents for non-distributed tasks, we evaluated the distributed spot training problem from the throughput, cost, and model size perspective on different hardware setups.
By including our insights, their technique for scheduling on spot instances could be adapted to optimize the total throughput of all peers.

Skypilot~\cite{yang2023skypilot} is a broker system where users can submit their hardware requirements, and it tries to provision the necessary resources on any supported cloud.
It features a preemption analysis that counts the number of interruptions in a zone and can decide to migrate whenever they cross a certain threshold.
We have shown that multi-, hybrid-cloud, and geo-distributed training is possible, and by combining our insights, it would open up auto-migrated, decentralized DL training for the best spot prices in the world.
\section{Conclusion}
\label{sec:conclusion}
This paper analyzes multi- and hybrid-cloud training in a decentralized fashion on spot instances. 
We define the lower bounds of model sizes that can be scaled cost-efficiently using the granularity metric to estimate their suitability for distributed training in low-bandwidth, high-latency situations.
We show that training on multiple cloud providers and four continents still scales with additional compute resources.
Alternatively to the current use of spot instances in DL, we show the potential of using spot instances in a distributed, decentralized way by being more cost-efficient with eight T4 instances over a DGX-2 from the same cloud provider while paying additional egress costs.
Finally, we provide an intuition about where costs in such a training scenario come from and how different model sizes from CV and NLP affect throughput and costs.
Our work empowers practitioners to utilize spot-priced instances for distributed deep learning with relatively small models. 
Our insights show some potential that can further improve distributed training performance, such as optimizers with higher minibatch sizes and improvements regarding the communication time with, e.g., better compression.
\section{Appendix: ASR Case Study}
\label{sec:appendix}
\begin{figure}
    \begin{subfigure}[c]{0.24\textwidth}
        \includegraphics[width=\textwidth]{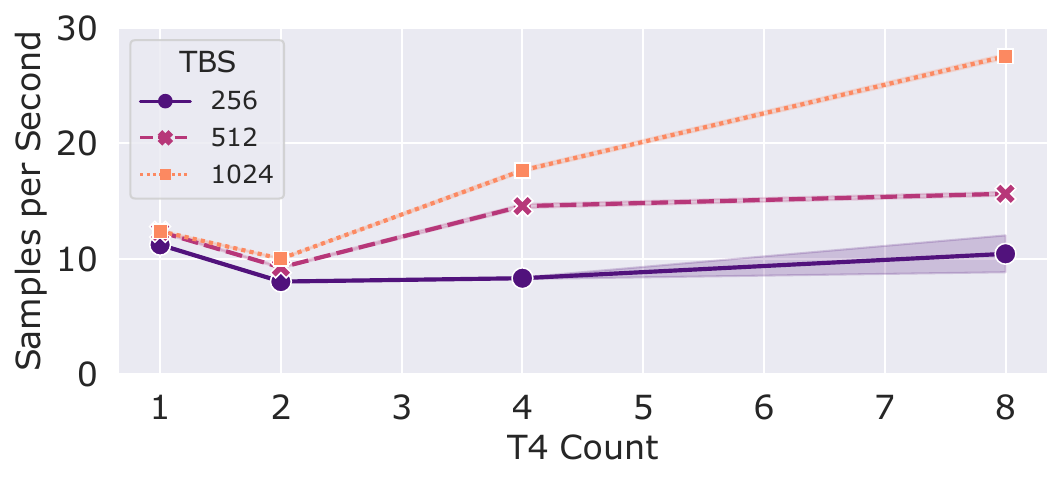}
        \vspace{-6mm}
        \caption{Throughput}
        \label{fig:asr-small-throughput}
    \end{subfigure}
    \begin{subfigure}[c]{0.21\textwidth}
        \includegraphics[width=\textwidth]{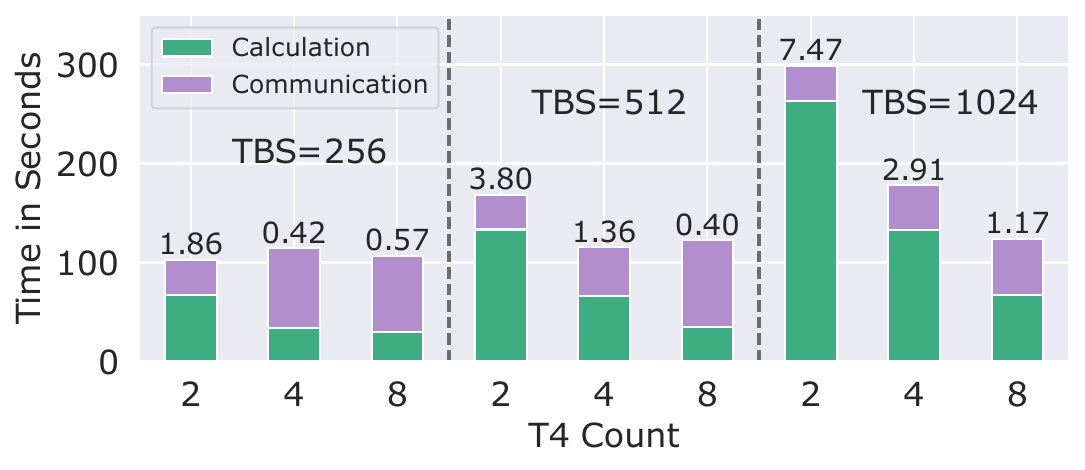}  
        \vspace{-4mm}
        \caption{Granuarity}
        \label{fig:asr-small-granularity}
    \end{subfigure}
    \vspace{-3mm}
    \caption{WhisperSmall performance with varying TBS.}
    \label{fig:asr-small-performance}
    \vspace*{-2mm}
\end{figure} 
\begin{figure}
    \includegraphics[width=0.45\textwidth]{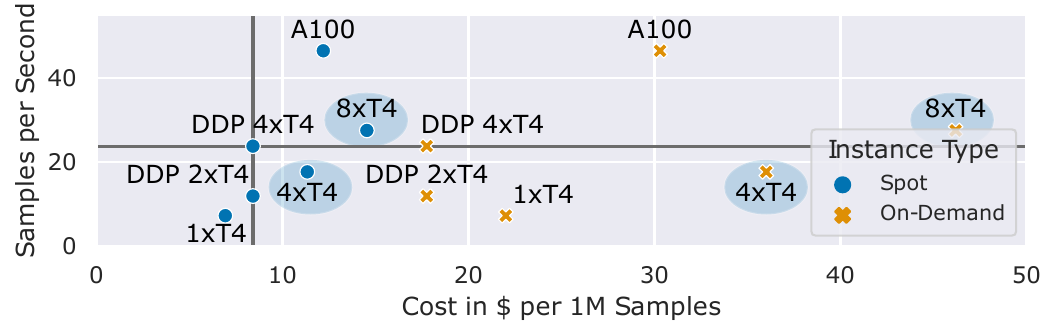}
    \vspace{-3mm}
    \caption{{\color{diff1}Cost to throughput tradeoff for WhisperSmall at TBS=1024 with different instance types. Our training setups (circled) provide mixed result of being slightly faster and more expensive than comparable, centralized DDP offering.}} 
    \label{fig:asr-sps-trade-off}
    \vspace*{-6mm}
\end{figure}

We perform a case study on Automatic Speech Recognition (ASR) to showcase spot training on weaker GPUs. 
Whisper~\cite{pmlr-v202-radford23a} is a state-of-the-art ASR model trained on 680,000 hours of labeled data to transcribe audio.
It features different sizes, from 37.8M to 1.5B parameters, and was trained with a minibatch size of 256.
We use the Commonvoice~\cite{ardila2019common} dataset, preprocessed to Log-Mel spectrograms.
In our distributed experiments, we start with a TBS of 256 and increase to 512 and 1024 to combat potential granularity issues.
Due to memory constraints, only three model sizes (Tiny, Base, Small) were trainable on the T4 GPU.
Unfortunately, the original TBS of 256 was not large enough to train the relatively small models due to their small granularity (0.04, 0.14 and 0.57 at 8xT4, respectively) with no performance benefits. 
The only model showing scaling potential is WhisperSmall, with a granularity of 1.8 with 2xT4.
However, when scaling the target batch size to 512 and 1024, we see some benefit over the single GPU runs for the WhisperSmall model (\Cref{fig:asr-small-performance}).
By effectively increasing the amount of computation by the factors of 2 and 4, we can generate a speedup of $1.27\times$ and $2.2\times$ with 8xT4's for the TBS 512 and 1024, respectively.
{\color{diff1}
When compared to other hardware setups, our A100 80GB GPU and the best multi-T4 GPU on GC (4xT4) with Pytorch DDP (\Cref{fig:asr-sps-trade-off}) have almost double the throughput at 46~SPS and are slightly slower at 24~SPS, respectively, compared to our 8xT4 setup which runs at 28~SPS.
This outcome is not surprising due to the generational leap in architecture for the A100 and the slower interconnect with our 8xT4 experiments compared to a single 4xT4 node (see \Cref{sec:model-suitability} for a detailed throughput analysis).
The proposed cost-throughput ratio is mixed: the A100 is at \$12.19/1M samples, the DDP 4xT4 is at \$8.41/1M, and our 8xT4 is at \$14.53/1M.
Our proposed setup is slightly more expensive than the A100, and it will not scale beyond eight T4 GPUs due a granularity at 1.17, leaving the A100 as the fastest and the DDP 4xT4 setup as the cheaper but slower alternative.
Despite these results, our proposed setup has several benefits, including resilience for spot interruptions, interruption-free migration to the lowest cloud prices, and the possibility to scale the GPU count up as long as granularity permits it.
}




\begin{acks}
This work is funded in part by the Deutsche Forschungsgemeinschaft (DFG, German Research Foundation) - 392214008.
\end{acks}

\bibliographystyle{ACM-Reference-Format}
\bibliography{main}

\end{document}